\documentclass[sigconf]{acmart}
\AtBeginDocument{%
  \providecommand\BibTeX{{%
    \normalfont B\kern-0.5em{\scshape i\kern-0.25em b}\kern-0.8em\TeX}}}

\setcopyright{acmcopyright}
\copyrightyear{2018}
\acmYear{2018}
\acmDOI{10.1145/1122445.1122456}

\acmConference[Woodstock '18]{Woodstock '18: ACM Symposium on Neural
  Gaze Detection}{June 03--05, 2018}{Woodstock, NY}
\acmBooktitle{Woodstock '18: ACM Symposium on Neural Gaze Detection,
  June 03--05, 2018, Woodstock, NY}
\acmPrice{15.00}
\acmISBN{978-1-4503-XXXX-X/18/06}

\usepackage{subcaption,booktabs}
\usepackage{multirow}
\usepackage{enumitem}
\usepackage{float}
\usepackage{xspace}
\usepackage{color}

\newcommand{\ie}{\emph{i.e.,}\xspace}
\newcommand{\eg}{\emph{e.g.,}\xspace}
\def\B#1{\mathbf #1}
\def\C#1{\mathcal #1}
\setlist[itemize]{itemsep=-0.25ex,leftmargin=2.5ex}
\setlist[enumerate]{itemsep=-0.25ex,leftmargin=2.5ex}
\begin{document}

\title{KECRS: Towards Knowledge-Enriched Conversational Recommendation System}

\author{\textbf{Tong Zhang$ ^1$, Yong Liu$ ^{2,3}$, Peixiang Zhong$ ^{2,3}$, Chen Zhang$ ^4$, Hao Wang$ ^4$, Chunyan Miao$ ^{1,2,3}$}}
\affiliation{
\institution{$^1$ School of Computer Science and Engineering, Nanyang Technological University (NTU), Singapore}
\country{}
}
\affiliation{
\institution{$^2$ Alibaba-NTU Singapore Joint Research Institute, NTU, Singapore}
\country{}
}
\affiliation{
\institution{$^3$ Joint NTU-UBC Research Centre of Excellence in Active Living for the Elderly, NTU, Singapore}
\country{}
}
\affiliation{
\institution{$^4$ Alibaba Group, China}
\country{}
}
\email{{tong24, stephenliu, peixiang001, ascymiao}@ntu.edu.sg, zhangchen010295@163.com, cashenry@126.com}
\renewcommand{\shortauthors}{Zhang et al.}

\begin{abstract}
The chit-chat-based conversational recommendation systems (CRS) provide item recommendations to users through natural language interactions. To better understand user's intentions, external knowledge graphs (KG) have been introduced into chit-chat-based CRS. However, existing chit-chat-based CRS usually generate repetitive item recommendations, and they cannot properly infuse knowledge from KG into CRS to generate informative responses. 
To remedy these issues, we first reformulate the conversational recommendation task to highlight that the recommended items should be new and possibly interested by users. Then, we propose the Knowledge-Enriched Conversational Recommendation System (KECRS). Specifically, we develop the Bag-of-Entity (BOE) loss and the infusion loss to better integrate KG with CRS for generating more diverse and informative responses. BOE loss provides an additional supervision signal to guide CRS to learn from both human-written utterances and KG. Infusion loss bridges the gap between the word embeddings and entity embeddings by minimizing distances of the same words in these two embeddings. Moreover, we facilitate our study by constructing a high-quality KG, \ie The Movie Domain Knowledge Graph (TMDKG). Experimental results on a large-scale dataset demonstrate that KECRS outperforms state-of-the-art chit-chat-based CRS, in terms of both recommendation accuracy and response generation quality.

\end{abstract}

\maketitle

\section{Introduction}
Recommendation system has been widely applied on e-commerce platforms to provide personalized services to customers~\cite{schafer2001commerce}. 
In general, the recommendation system matches the user's preferences with the items based on her historical behaviors, \eg clicking history, and rating history. Although existing recommendation methods can usually achieve satisfied recommendation accuracy, they still suffer from the cold-start problem, where no historical observation is available for new users~\cite{bobadilla2013recommender}. 
Moreover, user’s demands may also vary over time. Thus, it is very hard for traditional recommendation systems to capture user's dynamic interests timely~\cite{jannach2020survey}. 

The recently emerging conversational recommendation system (CRS) becomes an appealing solution to solve these two problems. 
A CRS can interact with users by natural language and obtain users' explicit feedback timely for better understanding users' dynamic interests. Previous studies about CRS can be classified into two main categories: 1) attribute-based CRS, and 2) chit-chat-based CRS~\cite{jannach2020survey}. 
The attribute-based conversational recommendation methods~\cite{lei2020estimation,lei2020interactive, sun2018conversational,christakopoulou2016towards, christakopoulou2018q, zhang2018towards, zou2020towards} mainly focus on the recommendation task. They explore the user's preferences on different item attributes to retrieve the items matching the user's interests. Their objective is to optimize the recommendation accuracy and the number of interaction rounds between the user and the recommender agent. Differing from these attribute-based methods, the chit-chat-based conversational recommendation methods~\cite{li2018towards, chen2019towards, zhou2020improving, sarkar2020suggest, liu2020towards, zhou2020towards, hayati2020inspired} naturally integrate the recommendation task and the response generation task. In real application scenarios, more natural and human-like dialogues may improve user's experiences and make users more engaged \cite{yan2017building}. Thus, we focus on the chit-chat-based CRS in this work. 

\begin{figure}[]
    \centering
    \includegraphics[width=0.975\columnwidth]{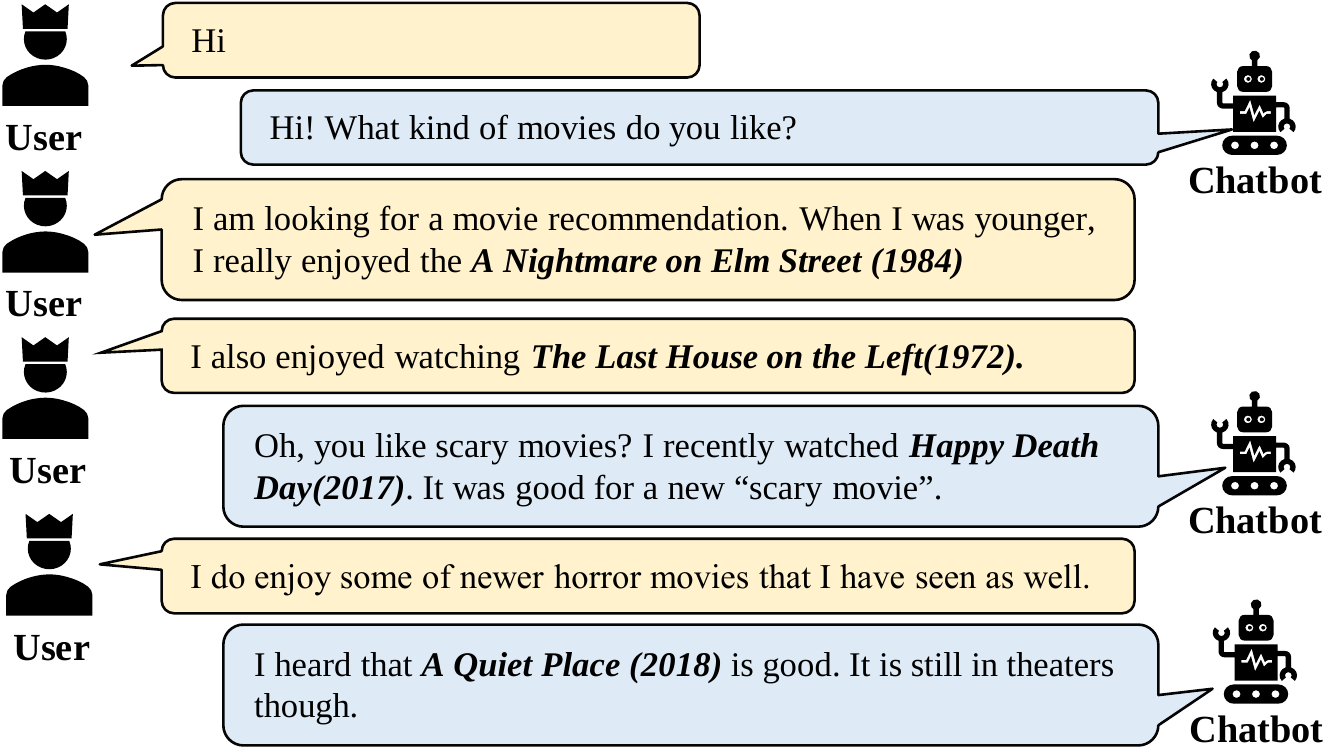}
    \caption{An example of a conversation between a user and the Chatbot for movie recommendation.}
    \label{fig:crs_example}
\end{figure}

Generally, a chit-chat-based CRS is composed of two main modules, \ie the recommendation module and the response generation module. The recommendation module firstly explores the user’s intentions and recommends a list of items that meet her interests. Then, the response generation module generates appropriate responses based on the recommended item list and the dialogue context, which includes the user's timely feedback. To support the research about chit-chat-based CRS, the pioneer work ~\cite{li2018towards} releases a large conversational recommendation dataset in the movie domain, namely \textit{REcommendations through DIALog (REDIAL)}. However, a conversation may only contain several utterances and lack sufficient contextual information. As an example conversation in Figure ~\ref{fig:crs_example}, if we have no prior knowledge that \textit{A Nightmare on Elm Street} and \textit{The Last House on the Left} are both thrillers, we cannot recommend another thriller (\eg \textit{Happy Death Day}) to the user. Besides, an ideal CRS should provide not only accurate recommendations but also informative responses that contain information about the recommended items (\eg \textit{a new "scary movie"}). Based on these concerns, Chen et al. \cite{chen2019towards} and Zhou et al. \cite{zhou2020improving} exploit the knowledge graph (KG) to improve the performances of CRS.

Although these studies have made progress in CRS, they still suffer from the following limitations:
\begin{itemize}[noitemsep,topsep=0pt,leftmargin=*]
    \item Previous studies such as~\cite{zhou2020improving, chen2019towards} treat all items mentioned by the recommender as recommendations. Thus, they focus on the item mention prediction task instead of the item recommendation task. 
    Under the settings of these studies, items may be repetitively recommended by the recommender agent, which may hurt the user's experiences.
    
    \item Previous works are more likely to generate dull and generic responses. To generate more diverse responses, 
    external KGs (\eg DBpedia and ConceptNet) are usually introduced into the CRS~\cite{chen2019towards, sarkar2020suggest, zhou2020improving}. 
    As the construction of existing conversational recommendation dataset is not based on the external KGs, there exists a big semantic gap between the information in conversation utterances and the knowledge in KGs. 
    For example, \cite{chen2019towards,sarkar2020suggest,zhou2020improving} use DBpedia subgraph as the external KG and REDIAL as the conversational recommendation dataset. Through our data analysis, REDIAL conversations only mention 8\% of entities in DBpedia subgraph excluding movie entities. At the same time, each response by the recommender in REDIAL only contains 0.25 entities, on average. In this case, KG entities are hard to be generated in responses by previous models. The effect of the KG in the response generation module may also be limited. Moreover, KG entities' embeddings are learned in the recommendation module while words embeddings are learned in the generation module \cite{chen2019towards, zhou2020improving, jannach2020survey}. The representation space gap between these two embeddings may also limit the capability of the models to generate KG entities in responses. However, existing methods mostly regard the feature obtained from the recommendation module just as an additional feature of the response generation module \cite{chen2019towards, zhou2020improving}. They do not model the relationship among different spaces explicitly.
    \item KGs used in the previous work \cite{chen2019towards, sarkar2020suggest, zhou2020improving} are noisy and incomplete. They mostly extract a subset of an open domain KG (\eg DBpedia and ConceptNet) and then use it in a specific domain (\eg movie domain). These KGs may contain irrelevant information and lose high-order neighbors, which limits their contributions to both the recommendation and response generation module.
\end{itemize} 

To this end, in this paper, we first reformulate the conversational recommendation task to highlight that recommended items need to be new and possibly interested in by users. Using KG as external knowledge, we propose the Knowledge-Enriched Conversational Recommendation System (KECRS), which can generate both more informative responses and more accurate recommendations. Specifically, we propose Bag-of-Entity (BOE) loss and infusion loss to generate more diverse and informative responses. BOE loss provides an additional supervision signal to guide the model to generate knowledge-enriched responses not only from utterances in training data but also from the neighboring entities of recommended items in KG. Infusion loss bridges the representation space gap between words in the generation module and entities in the recommendation modules by minimizing distances of the same entities in different spaces. With these two losses, entities related to recommended items are more likely to be generated in responses. Thus, the diversity and informativeness of responses can be highly improved. Moreover, we facilitate our study by constructing a high-quality KG, The Movie Domain Knowledge Graph (TMDKG) related to the conversational recommendation dataset REDIAL \cite{li2018towards}. Extensive experiments demonstrate that the proposed two losses help improve model performance and the proposed KECRS model outperforms state-of-the-art CRS models on a large-scale public dataset in terms of both recommendation accuracy and response quality. In addition, TMDKG is able to improve the performances of both the KECRS model and baseline methods.

\begin{figure*}[!t]
    \centering
    \includegraphics[width=\textwidth]{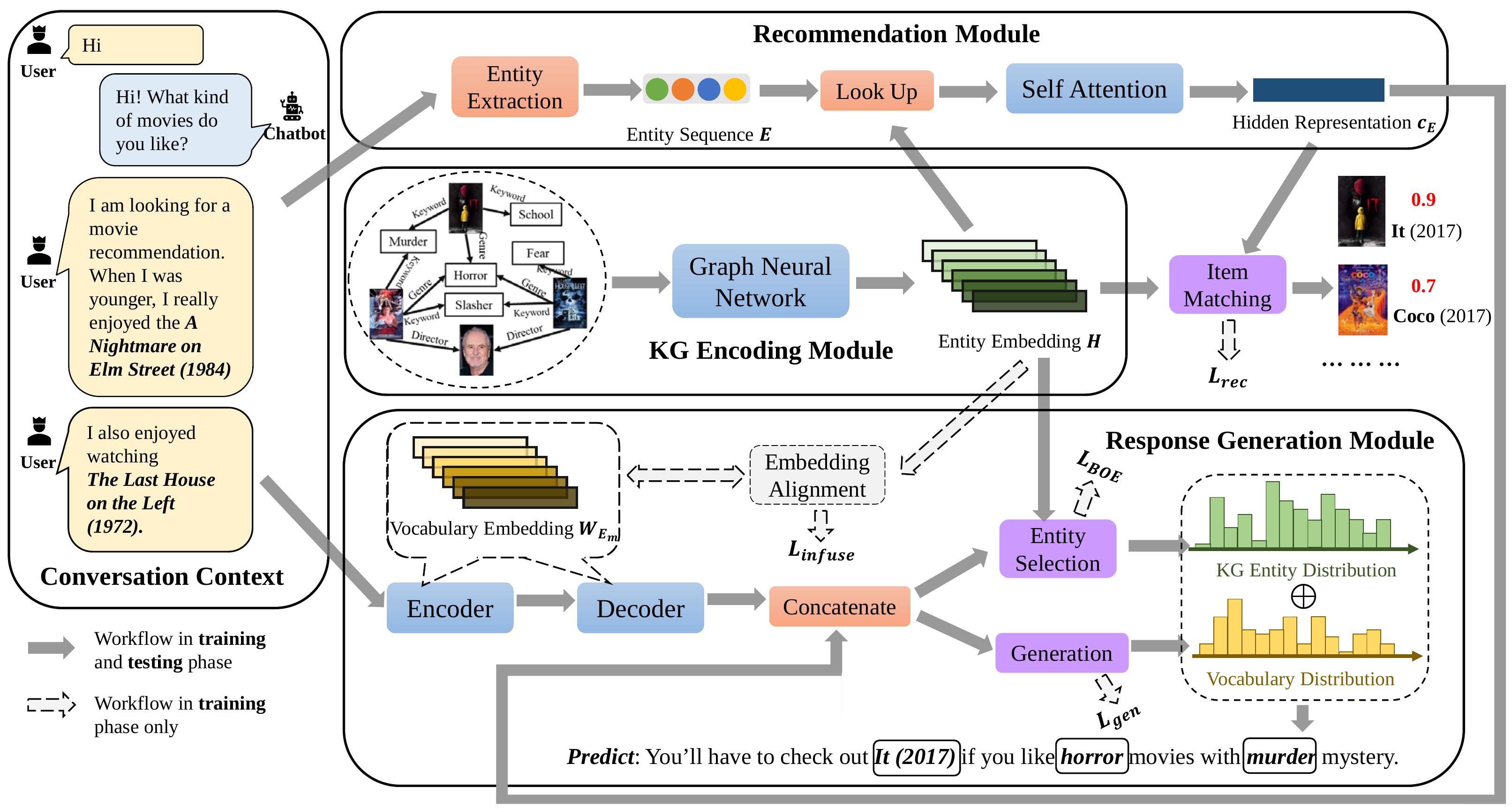}
    \caption{The overall framework of the proposed KECRS model.}
    \label{fig:my_label}
\end{figure*}

\section{Related work}
Traditional recommendation systems may have cold-start problems for new users and be hard to capture users' dynamic preferences. To tackle these problems, different methods have been proposed, \eg introducing external knowledge \cite{zhang2016collaborative, liu2020heterogeneous} and analyzing review data \cite{liu2020neural, ji2019recommendation}. 
However, these methods are mostly static or have limited assumptions \cite{gao2021advances}.
With the developments of dialogue systems in open-domain \cite{zhong2019affect, dinan2018wizard, koncel2019text} and task-oriented \cite{madotto2018mem2seq, wu2019global, ren2020crsal} tasks, conversational recommendation systems become an appealing solution to capture users' dynamic preferences.

One category of conversational recommendation systems are attribute-based conversational recommendation systems~\cite{lei2020estimation, lei2020interactive, sun2018conversational, christakopoulou2016towards, christakopoulou2018q, zhang2018towards, zou2020towards}. Most studies in this category focus on how to provide high-quality recommendations within the shortest number of conversation turns but do not pay much attention to generate human-like responses. Christakopoulou et al. \cite{christakopoulou2016towards} first propose a contextual bandit-based method to develop the conversational recommendation system and then propose an RNN-based method \cite{christakopoulou2018q}, which jointly optimize attribute and item prediction tasks. But these studies restrict the conversation to two turns, \ie, one turn for asking and the other for recommending. Sun and Zhang \cite{sun2018conversational} extend it to the multi-turn conversations but still remain the single-round recommendation setting. Following this setting, Zhang et al.\cite{zhang2018towards} propose the Multi-Memory Network architecture, which conducts question prediction and search in a parallel manner. However, in practical deployments, users may reject a recommendation sometimes, which is ignored by the previous single-round recommendation setting. Thus, Lei et al. \cite{lei2020estimation, lei2020interactive} adopt the multi-round setting, where a conversational recommendation system can conduct multi-turn conversations and do recommendations multi-times.

The other category of conversational recommendation methods are chit-chat-based conversational recommendation systems~\cite{li2018towards, chen2019towards, zhou2020improving, sarkar2020suggest, liu2020towards, zhou2020towards, hayati2020inspired}. Most studies in this category focus on both giving accurate recommendations and generating natural and human-like responses. As there is no publicly available large-scale dataset consisting of real-world dialogues centered around recommendations, Li et al. \cite{li2018towards} release a conversational recommendation dataset in the movie domain and propose an HRED-based \cite{serban2016hierarchical} baseline model. As it is hard to understand user's intentions only from utterances, Chen et al.~\cite{chen2019towards} introduce KG into Chit-chat-based CRS and propose a knowledge-based recommender dialog system. Based on~\cite{chen2019towards}, Sarkar et al.~\cite{sarkar2020suggest} explore the performances of using different sizes of KGs in the recommendation module. To better understand the user's preferences, Zhou et al. \cite{zhou2020improving} leverage the entity-oriented KG (\ie DBpedia) and the word-oriented KG (\ie ConceptNet). To make the recommendation proactively and naturally, Liu et al. \cite{liu2020towards} and Zhou et al.~\cite{zhou2020towards} use topics to guide dialogue from non-recommendation to recommendation and propose the topic-guided conversational recommendation task.

\section{Preliminary}
The conversational recommendation task studied in this work consists of two sub-tasks, \ie the recommendation task and the response generation task. The objective of the recommendation task is to recommend items that are new and possibly interested in by users. However, previous works~\cite{zhou2020improving, chen2019towards} consider all items mentioned by recommenders as recommendations. This setting will cause the following two issues. 
Firstly, the recommendation models are more likely to generate repetitive items, which may hurt user's experiences. As shown in Table ~\ref{top1_table}, almost half of the Top-1 recommended items of KBRD \cite{chen2019towards} and KGSF~\cite{zhou2020improving} have been mentioned or recommended before. However, this ratio in the REDIAL dataset is only 8.5\%. Thus, if we do not constrain the recommendation setting, repetitive items may keep being recommended, which may highly hurt user's experiences. Secondly, it is unreasonable to treat the items disliked by the user as correct item recommendations. 
In the experiments, around 15\% of items hit by KBRD and KGSF at Top-1 are not liked by users. If we treat them as correct recommendations, it conflicts with the objective of recommendation systems and makes evaluation metrics higher than practical effect. For the response generation task, it aims to generate responses given the dialogue contexts and the recommended item list.

Following~\cite{chen2019towards, sarkar2020suggest, zhou2020improving}, we also introduce the KG as the external knowledge to understand the user's preferences and generate more informative responses. However, KGs used in the previous studies~\cite{chen2019towards, sarkar2020suggest, zhou2020improving} are mostly the sub-graph of open-domain KGs, \eg DBpedia and ConceptNet, which may contain irrelevant information and lose high-order neighbors of an entity. To alleviate these issues, we build a high-quality KG, \ie TMDKG, in the movie domain. TMDKG is constructed on the information from The Movie Database (TMDb)\footnote{https://www.themoviedb.org/}, which is a community built movie and TV database. More details about TMDKG can be found in appendix~\ref{TMDKG Appendix}.

Formally, Let $X = \left\{ x_1, x_2,...,x_n\right\}$ denotes the utterances of a conversation, where $n$ denotes the number of conversation histories. Let $\mathcal{G} = \{(v_h, r, v_t)|v_h,$ $v_t \in V, r \in R\}$ denote the KG, where each triplet $(v_h, r, v_t)$ describes that there is a relationship $r$ between the head entity $v_h$ and the tail entity $v_t$. The conversational recommendation task can be formulated as learning two functions $f(X, \mathcal{G})$ and $g(X, \mathcal{G}, f(X, \mathcal{G}))$. Here, the function $f(X, \mathcal{G})$ predicts \textit{\textbf{new}} and user possibly \textit{\textbf{interesting}} items based on contexts $X$ and the KG $\mathcal{G}$, and the function $g(X, \mathcal{G}, f(X, \mathcal{G}))$ generates human-like responses given contexts $X$, the KG $\mathcal{G}$, and the list of items recommended by $f(X, \mathcal{G})$. During the inference phase, both $f$ and $g$ are integrated to generate a natural language response correlated to the context.

\begin{table}[]
    \centering
    \caption{Statistics of Top-1 recommended items of KBRD \cite{chen2019towards} and KGSF \cite{zhou2020improving} and items mentioned by the recommender in the REDIAL \cite{li2018towards} dataset.}
    \begin{tabular}{c|c|c|c|c}
        \hline
         \multirow{2}*{Model} & \multicolumn{2}{c|}{Items recommended at Top-1} &\multicolumn{2}{c}{Items in REDIAL} \\
         \cline{2-5}
         & New & Repetitive & New & Repetitive\\
         \hline
         KBRD & 46.0\% & 54.0\% &\multirow{2}*{91.5\%} & \multirow{2}*{8.5\%}\\
         \cline{1-3}
         KGSF & 55.9\% & 44.1\% & &\\
         \hline
    \end{tabular}
    \label{top1_table}
\end{table}





\section{Model Structure}
In this section, we present the proposed Knowledge-Enriched Conversational Recommendation System (KECRS). As shown in Figure \ref{fig:my_label}, KECRS consists of three main components: 1) knowledge graph encoding module, 2) recommendation module, and 3) response generation module. Next, we introduce the details of each component.


\subsection{Knowledge Graph Encoding Module}

As relation semantics are important to model the similarity of two movies, we adopt R-GCN \cite{schlichtkrull2018modeling} to embed the structural and relational information in $\mathcal{G}$ to learn entity representations. Specifically, at the $(l+1)$-th layer, the representation of an entity $i$ in $\mathcal{G}$ is defined as,
\begin{equation}
    \mathbf{h}_i^{(l+1)} = \sigma\bigg(\sum_{r \in R}\sum_{j \in N_i^r}\frac{1}{c_{i,r}}\mathbf{W}_r^{(l)}\mathbf{h}_j^{(l)} + \mathbf{W}_0^{(l)}\mathbf{h}_i^{(l)}\bigg),
\end{equation}
where $\mathbf{h}_i^{(l+1)} \in \mathbb{R}^{d_k}$ denotes the embedding of the entity $i$ at $(l+1)$-th layer, and $d_k$ denotes the dimension of the entity embedding at $(l+1)$-th layer. $N_i^r$ denotes the neighbor set of entity $i$ under relation $r \in R$. $\mathbf{W_r^{(l)}}$ is a learnable relation-specific transformation matrix for the embedding of neighboring nodes under relation $r$, and $\mathbf{W_0^{(l)}}$ is a learnable transformation matrix for the self-connection relation of the embedding of entity $i$. $c_{i,r}$ is the normalization constant which is $|N_i^r|$ here.

After perform $L$ layers, we obtain multiple representations for each entity $i$, namely $\{ \mathbf{h}_i^{(0)}, \mathbf{h}_i^{(1)}, ..., \mathbf{h}_i^{(L)}\}$. To integrate different depth information, we apply the layer-aggregation mechanism proposed in~\cite{xu2018representation} to compute the representation $\mathbf{h}_i$ as follows,
\begin{equation}
    \mathbf{h}_i = \mathbf{W}_h([\mathbf{h}_i^{(0)};\mathbf{h}_i^{(1)}; ...;\mathbf{h}_i^{(L)}]) + \mathbf{b}_h,
\end{equation}
where [;] denotes the concatenation operation, $\mathbf{W}_h \in \mathbb{R}^{d_f \cdot (L*d_k)}$ and $\mathbf{b}_h \in \mathbb{R}^{d_f}$ are learnable parameters. Finally, we can obtain the hidden representations of all entities in $\C{G}$, which is denoted by $\mathbf{H} \in \mathbb{R}^{(|V| \cdot d_f)}$.
    
\subsection{Recommendation Module}

To exploit KG for understanding the user's intentions, we firstly map the conversation context $X$ to an entity sequence $E$ by simply checking each entity in the KG. After looking up the hidden representations of entities in $E$ from $\mathbf{H}$, we can obtain the matrix of entities' hidden representation $\mathbf{H}_E \in \mathbb{R}^{|E| \cdot d_f}$. Then, we apply the self-attentive mechanism \cite{chen2019towards, zhou2020improving} on $\mathbf{H}_E$ to obtain the hidden representation $\mathbf{c}_E$ of $E$ as follows,
\begin{equation}
\begin{split}
    &\mathbf{c}_E = \mathbf{\alpha} \mathbf{H}_E,\\
    & \mathbf{\alpha} = \mbox{softmax}\big(\mathbf{W}_k\mbox{tanh}(\mathbf{W}_q\mathbf{H}_E^\top)\big),
\end{split}
\end{equation}
where $\mathbf{W}_q\in\mathbb{R}^{d_q\cdot d_f}$ and $\mathbf{W}_k \in \mathbb{R}^{d_q}$ are learnable parameters. $\mathbf{\alpha} \in \mathbb{R}^{|E|}$ is the importance vector of entities in the sequence E.

To compute the probability over item candidates, we conduct inner product between the sequence representation $\mathbf{c}_E$ and the item representation $\mathbf{H}_I$ as,
\begin{equation}
    P_{rec} = \mbox{softmax}(\mathbf{c}_E\mathbf{H}_I^\top),
\end{equation}
where $\mathbf{H}_I$ is the matrix selecting from $\mathbf{H}$ that only contains the hidden representation of items.

To optimize the recommendation module, we minimize the cross-entropy loss for selecting the true items from a list of items. Specifically, the loss $L_{rec}$ is defined as follows,
\begin{equation}
    L_{rec} = -\mathbb{E}_{m \in M}\bigg[\mathbb{E}_{(E, y_E) \in m}\big[\sum_{i=1}^{N}y_E\log (P_{rec}^{(E)}(i))\big]\bigg],
\end{equation}
where $M$ denotes the set of conversations, $E$ denotes the sequence of entities, and $y_E$ denotes the item label. $N$ is the number of item candidates. To meet the traditional recommendation setting, we only remain the case that has new and user-liked recommendations. 

\subsection{Response Generation Module}

The response generation module aims to generate appropriate responses to the user with the given conversation context. As Transformer\cite{vaswani2017attention} has shown its excellent performances in the language representation learning~\cite{devlin2018bert}, we leverage Transformer to develop an Encoder-Decoder framework for the response generation task, considering conversation context utterances $X$, sequence representation $\B{c_E}$ from recommendation module, and entity representation $\B{H}$.

\subsubsection{Encoder}
Given the context utterances $X$, we first concatenate them as a single context sentence $X_{all}$ by a special token `\_split\_'. Then, we use an embedding layer $E_m$ to convert each word $t_i$ in $X_{all}$ into a vector $\mathbf{t}_i \in \mathbb{R}^{d_t}$, as such, the conversation context $X_{all}$ is represented as follows,
\begin{equation}
        T = \mathbf{\{t}_0, \mathbf{t}_1,\cdots, \mathbf{t}_{\hat{n}}\},
\end{equation}
where $\hat{n}$ is the length of the context sequence. After feeding $T$ into $n_e$ Transformer encoder layers, we obtain the feature vector $\mathbf{t}_i^{n_e}$ of each word $i$ in the context sentence as follows,
\begin{equation}
    T_{n_e} = \mathbf{\{t}_0^{(n_e)}, \mathbf{t}_1^{(n_e)}, \cdots, \mathbf{t}_{\hat{n}}^{(n_e)}\}.
\end{equation}

\subsubsection{Decoder}
\label{ss:decoder}
At each decoding time step $j$, we feed $T_{n_e}$ and the ground truth word sequence before $j$ into $n_d$ Transformer decoder layers and obtain a hidden representation vector $\mathbf{s}_j \in \mathbb{R}^{d_{res}}$. Thus, the probability distribution over the vocabulary $V_{oc}$ at decoding time step $j$ is computed as follows,
\begin{equation}
    P_{res} = \mbox{softmax}\big(\phi([\mathbf{s}_j;\mathbf{c}_E])\mathbf{W}_{E_m}^T + \mathbf{b}_{res}\big),
\label{equation_ref}
\end{equation}
where $\mathbf{c}_E \in \mathbb{R}^{d_f}$ is the knowledge-enriched representation obtained from the recommendation module, $\mathbf{W}_{E_m} \in \mathbb{R}^{|V_{oc}| \cdot d_{res}}$ is the weight matrix of the embedding layer $E_m$, $\phi(\cdot)$ is the linear function to align the dimension, and $\mathbf{b}_{res}$ is the bias vector. To learn the response generation module from the ground truth responses, we define the following generation loss,
\begin{equation}
    L_{gen} = -\mathbb{E}_{m\in M}[\mathbb{E}_{(X, Y) \in m}[\frac{1}{n_y}\sum_{j=1}^{n_y}\log(P_{res}(y_j|X, y_0, y_1, ..., y_{j-1}))]],
\end{equation}
where $M$ denotes the set of conversations, $X$ denotes the context utterance, $Y$ denotes the ground-truth response, and $n_y$ denotes the sentence length of $Y$.

\subsubsection{Entity Selection}

Different from traditional chit-chat dialogue systems, the response generator of a CRS is expected to contain both recommended items and relevant information. Thus, it is important to infuse the knowledge from KG into the response generation. However, as there is little related information about recommended items in the conversation dataset, it is hard for the model to learn to generate informative sentences. Moreover, sentences with more relevant information but not in the training data may be punished even more than generic sentences (\eg I don't know). 

Inspired by \cite{ma2018bag,zhao2017learning}, we introduce the Bag-of-Entity (BOE) loss to cope with this issue. The intuition behind BOE loss is that BOE loss can provide an additional supervision signal, and thus guide the model to generate knowledge-enriched responses not only from utterances in training data but also from the neighboring entities of recommended items in KG. Moreover, as entities in KG do not have positional information, we design the BOE loss at the sentence-level instead of the word-level. Specifically, at each decoding time step $j$, we compute a probability distribution over all entities in the KG as,
\begin{equation}
    P_{res}^{'} = \mbox{softmax}\big([\mathbf{s}_j;\mathbf{c}_E]\mathbf{W}_{align}\mathbf{H}^\top + \mathbf{b}_{res}^{'}\big),
\end{equation}
where $\mathbf{s}_j$ and $\mathbf{c}_E$ are the same as the ones used in Eq.~\eqref{equation_ref}. $\mathbf{W}_{align} \in \mathbb{R}^{(d_{res}+d_f)\cdot d_{f}}$ is the weight matrix, and $\mathbf{H}$ is the embedding matrix of all entities in the KG.

Then, we sum up the word-level score at each time step and obtain the sentence-level score. After normalizing the sentence-level score with the sigmoid function, we obtain the probability for each entity in the KG occurring in responses. Thus, the sentence-level score of Bag-of-Entity can be written as follows,
\begin{equation}
    P_{boe} = \mbox{sigmoid}(\sum_{j=1}^{L}P_{res}^{'}(y_j|X, y_0, y_1, ..., y_{j-1})),
\end{equation}
where $L$ is the length of generated sentence, $X$ denotes the context utterance, and $y_j$ denotes the j-th word in ground truth response.

Here, the target entities are the one-hop neighbors of recommended items in the KG. Thus, the BOE loss is defined as follows,
\begin{equation}
    L_{BOE} = -\mathbb{E}_{m \in M}\big[\mathbb{E}_{(X, y_{1-hop}) \in m}[\sum_{i=1}^N y_{1-hop}\log(P_{boe}^{(X)}(i))]\big],
\end{equation}
where $M$ denotes the set of conversations, $X$ denotes the context utterance, $N$ denotes the number of entities in the KG, and $y_{1-hop}$ denotes the label of the recommended items' one-hop neighbors in the KG.

\subsubsection{Embedding Alignment}

In practice, there usually exists a gap between the representation space of words in the generation module and the representation space of entities in the recommendation module. To align the embeddings in these two spaces, we propose the infusion loss. The core idea of infusion loss is to minimize the distance between the same word in these two hidden spaces. As a conversation may contain a large number of words, it is very time-consuming to enumerate all the word-entity pairs. Thus, we use the hidden representation $\mathbf{c}_E$ to represent the entity sequence of the conversation and calculate the similarity between $\B{c_E}$ and the word embedding $\mathbf{W}_{E_m}$ of the vocabulary $V_{oc}$ as follows,
\begin{equation}
    \mathbf{S}_{imlarity} = \phi^{'}(\mathbf{c}_E)\mathbf{W}_{E_m}^T + \mathbf{b}_{c_E},
\end{equation}
where $\phi^{'}(\cdot)$ is the linear function used to align the dimension of $\mathbf{c}_E$ and $\mathbf{W}_{E_m}$, and $\mathbf{b}_{c_E} \in \mathbb{R}^{|V_{oc}|}$ is a learnable bias vector. Then, the infusion Loss can be defined as follows,
\begin{equation}
    L_{infuse} = -\mathbb{E}_{m \in M}\big[\mathbb{E}_{(E, d_E)\in m} \big(\big\|\mathbf{S}_{imarity} - \mathbf{d}_E\big\|\big)\big],
\end{equation}
where the $M$ denotes the set of conversation, $E$ denotes the sequence of entities, and $d_E \in \{0,1\}^{|V_{oc}|}$ denotes the real distribution of entities in the vocabulary. Specifically, if an entity $e$ exits in $E$, the value of $\B{d_E}$ at the index of $e$ is 1, otherwise is 0. $\|\cdot\|$ denotes the $L_2$ distance between two vectors. 

Finally, to learn the parameters of generation module, we minimize the following objective function:
\begin{equation}
    L_{gen-all} = L_{gen} + \lambda_1 L_{BOE} + \lambda_2 L_{infuse},
\end{equation}
where $\lambda_1$ and $\lambda_2$ are two hyper-parameters that can be selected by cross-validation. In testing procedure, the probability distribution over the vocabulary at time step $j$ is calculated as follows,
\begin{equation}
    P_{all} = P_{res}(y_j|x, y_0^p, y_1^p, ..., y_{j-1}^p) + \lambda_3 P_{boe}(y_j|x, y_0^p, y_1^p, ..., y_{j-1}^p),
\end{equation}
where $x$ denotes the context utterance, $y_0^p, y_1^p, ..., y_{j-1}^p$ denotes the predicted sequence before time step $j$, and $\lambda_3$ is a hyper-parameter.

\section{Experimental Settings}
\label{experiment_section}
\begin{table*}[]
    \centering
    \caption{Recommendation performances of different method based on different knowledge graphs. $^{*}$ indicates that the improvement over the best baseline method is statistically significant with $p < 0.01$ using student \textit{t-test}.}
    \begin{tabular}{c|c|c|c|c|c|c|c|c|c|c|c|c}
    \hline
        \multirow{2}*{Model}& \multirow{2}*{KG} & \multicolumn{4}{c|}{Recall@$K$ (\%)} & \multicolumn{4}{c|}{Precision@$K$ (\%)} & \multicolumn{3}{c}{NDCG@$K$ (\%)} \\
    \cline{3-13}
        & & $K=1$ & $K=5$ & $K=10$ & $K=50$ & $K=1$ & $K=5$ & $K=10$ & $K=50$ & $K=5$ & $K=10$ & $K=50$ \\
    \hline
        REDIAL & - & 0.41 & 7.04 & 11.69 & 27.03 & 0.55 & 1.82 & 1.51 & 0.70 & 3.73 & 5.25 & 8.60 \\
    \hline
        \multirow{2}*{KBRD} & DBpedia & 1.40 & 8.52 & 13.54 & 30.12 & 2.03 & 2.14 & 1.78 & 0.78 & 5.06 & 6.89 & 10.61\\
         & TMDKG & 2.15 & 9.24 & 14.87 & 35.63 & 2.33 & 2.24 & 1.82 & 0.85 & 5.26 & 7.54 & 11.94 \\
    \hline     
        \multirow{2}*{COLING20} & DBpedia & 1.40 & 8.73 & 13.53 & 31.84 & 1.75 & 2.11 & 1.85 & 0.80 & 5.03 & 6.94 & 10.75 \\
         & TMDKG & 2.09 & 9.31 & 14.93 & 35.38 & 2.29 & 2.31 & 1.83 & 0.81 & 5.19 & 7.61 & 12.01 \\
    \hline
        \multirow{2}*{KGSF}& DBpedia& 1.91 & 9.19 & 13.42 & 32.76 & 1.96 & 1.90 & 1.49 & 0.68 & 5.64 & 7.38 & 11.57 \\
         & TMDKG& 2.13 & 7.82 & 13.68 & 34.87 & 2.13 & 1.56 & 1.35 & 0.70 & 5.01 & 6.82 & 11.59 \\
    \hline
        \multirow{2}*{KECRS} & DBpedia & 1.52 & 8.69 & 13.63  &  31.25 & 2.11 & 2.19 & 1.78  &  0.81 & 5.11 & 6.91  & 10.78  \\
         & TMDKG& \textbf{2.25}$^*$ & \textbf{9.45}$^*$ & \textbf{15.66}$^*$ & \textbf{36.61}$^*$ & \textbf{2.77}$^*$ & \textbf{2.39}$^*$ & \textbf{1.99}$^*$ & \textbf{0.90}$^*$ & \textbf{5.74}$^*$ & \textbf{7.92}$^*$ & \textbf{12.38}$^*$\\
    \hline     
    \end{tabular}
    \label{rec_ablation_table}
\end{table*}

In this section, we introduce the experimental datasets, evaluation metrics, baseline methods, and the model implementation details.

\subsection{Experimental Datasets}
The REDIAL dataset~\cite{li2018towards} is used as the conversational recommendation dataset for the experimental evaluation. This dataset is built through Amazon Mechanical Turk (AMT). Following a series of comprehensive instructions, the AMT workers generate dialogues centered around the movie recommendation. In REDIAL dataset, there are 10,006 conversations consisting of 182,150 utterances related to 51,699 movies. Following \cite{chen2019towards, zhou2020improving, li2018towards}, we split the dataset into training, validation, and testing sets using the ratio of 8:1:1.

For KG, previous works~\cite{chen2019towards, sarkar2020suggest, zhou2020improving} use the subgraph of an open domain KG, \ie DBpedia~\cite{dbpedia}, for improving the performances of CRS. The DBpedia used in this paper contains 64368 entities and 205650 edges. Our newly constructed TMDKG contains 15822 entities and 141564 edges.


\subsection{Evaluation Metrics}

For the recommendation task, we adopt precision, recall, and normalized discounted cumulative gain (NDCG) to evaluate the performances of top-$K$ item recommendation (respectively denoted by Recall@$K$, Precision@$K$, and NDCG@$K$). In the experiments, $K$ is empirically set to 1, 5, 10, and 50. As users may not be recommended too many movies in each conversation turn, Precision@1 and Precision@5 are more indicative of recommendation performances. 

Following \cite{chen2019towards,zhou2020improving}, we use Distinct n-gram (n=2, 3, 4) to measure the diversity of the generated responses. To automatically evaluate the informativeness of the generated responses, we introduce a new evaluation metric, named Average Entity Number (AEN), which denotes the average number of entities in each generated response. As ground-truth responses are not informative enough, we do not use metrics (\eg BLEU) to measure the similarity between ground-truth and generated utterances, which is also instructed by~\cite{zhou2020improving,jannach2020}. Instead, we adopt human evaluation to measure these aspects. we randomly sample 100 multi-turn conversations from the test set and invite three annotators to score responses generated by different models from the following aspects: 1) \textbf{Fluency}: whether responses are fluent; 2) \textbf{Relevancy}: whether responses are correlated with contexts; 3) \textbf{Informativeness}: whether responses contain rich information of recommended items. Each aspect is rated in $[0, 3]$, and final scores are the average of all annotators. For these evaluation metrics, the higher value indicates better performances.


\begin{table*}[]
    \centering
    \caption{Response generation performances of different methods.
    $^{*}$ indicates the improvement over the best baseline method is statistically significant with $p < 0.01$ using student \textit{t-test}. AEN is short for average entity number, which denotes the average number of entity in each generated response. Dist-2,3,4 is short for Distinct-2,3,4. All human evalution scores are from 0-3.}
	\begin{tabular}{c|c|c|c|c|c|c|c}
    \hline
        \multirow{2}*{Model} & \multicolumn{4}{c}{Automatic Evaluation}& \multicolumn{3}{|c}{Human Evaluation}\\
	\cline{2-8}
		& Dist-2 & Dist-3 & Dist-4 & AEN & Fluency & Relevancy & Informativeness \\
    \hline
        REDIAL & 0.10 & 0.18 & 0.24 & 0.08 & 1.92  & 1.62 & 1.05\\
    \hline 
        Transformer & 0.15 & 0.31 & 0.46 & 0.15 & 2.03  & 1.73 & 1.36\\
    \hline 
        KBRD & 0.26 & 0.30 & 0.45 & 0.15 & 2.10 & 1.72 & 1.32\\
    \hline 
        KGSF & 0.33 & 0.49 & 0.61 & 0.17 & 2.32 & 2.11 & 1.56\\
    \hline 
        KECRS & \textbf{0.48}$^*$ & \textbf{0.91}$^*$ & \textbf{1.23}$^*$ & \textbf{0.34}$^*$ & \textbf{2.56}$^*$ & \textbf{2.29}$^*$ & \textbf{2.18}$^*$\\
    \hline     
    \end{tabular}
    \label{response_eval_table}
\end{table*}

\subsection{Baseline Models}
We compare KECRS with the following baseline methods:


\begin{itemize}
    \item \noindent\textbf{REDIAL}~\cite{li2018towards}: This is a basic CRS, which is based on hierarchical recurrent encoder-decoder (HRED) architecture \cite{serban2016hierarchical}.
    
    \item \textbf{Transformer}~\cite{vaswani2017attention}: This is a basic transformer model that generates responses from utterance text only and does not contain a separate recommendation module.
    
    \item \textbf{KBRD}~\cite{chen2019towards}:This is a  knowledge-based CRS that employs DBpedia to understand the user's intentions. The response generation module is based on Transformer, where the KG information serves as a bias for generation.
    
    \item \textbf{COLING20}~\cite{sarkar2020suggest}: This model is based on KBRD. It studies the influences of the size of DBpedia on the performances of CRS. Here, we use the 5-hop subgraph of DBpedia which achieves the best results in the original paper. Note that this method only has a recommendation module.
    
    \item \textbf{KGSF}~\cite{zhou2020improving}: This knowledge-based CRS exploits both entity-oriented and word-oriented KGs (\ie DBpedia and ConceptNet) to enrich the data representations. It also introduces Mutual Information Maximization (MIM) to infuse the semantic space between two knowledge graph. The dialogue generation module in KGSF is also based on Transformer, where two KG-enriched decoders are adopted.
\end{itemize}


\subsection{Implementation Details}
We implement KECRS by PyTorch \cite{pytorch19}. For the recommendation module, we set the dimensionality of entity embedding $d_f$ and $d_k$ to 200. The number of R-GCN layers is set to 2. For response generation module, we set the dimensionality of word embeddings $d_t$ and the dimensionality of hidden representations $d_{res}$ of the transformer to 300. The number of transformer encoder and decoder layers is set to 2. The maximum length of transformer input $\hat{n}$ is set to 256. In the generation process, the maximum length of transformer output $n_y$ is set to 20. For model training, we use Adam optimizer~\cite{kingma2014adam} and set $\beta_1$ to 0.9, $\beta_2$ to 0.99, and $\epsilon$ to $1\times10^{-8}$. The learning rate is set to $3\times10^{-3}$ for recommendation module and $1\times10^{-1}$ for response generation module. The batch size is set to $128$ for the recommendation module and $32$ for the response generation module. Moreover, we also add the L2 regularization to avoid overfitting issues and apply gradient clipping to restrict the gradients within $[0, 0.1]$. The hyper-parameters $\lambda_1$, $\lambda_2$, and $\lambda_3$ are empirically set to $1.5$, $0.025$ and $0.1$ respectively.

\section{Experiments Results and Analysis}
In this section, we present the details about the experimental results.

\subsection{Recommendation Performances}
\label{ss:recperformances}

Table ~\ref{rec_ablation_table} summarizes the recommendation performances achieved by different methods. As shown in Table ~\ref{rec_ablation_table}, KBRD, COLING20, KGSF, and KECRS outperform REDIAL, because they introduce external KGs to understand the user's intentions. By using larger subgraph of DBpedia, COLING20 achieves better performances than KBRD. KGSF usually performs better than KBRD in terms of recall and NDCG, by incorporating a word-oriented KG (\ie ConcepNet) and an entity-oriented KG (\ie DBpedia) to enrich the entity representations. With the knowledge from DBpedia, KECRS achieves better Precision@1 than baseline methods. This indicates that the proposed KECRS model can achieve comparable recommendation performances with baseline methods, based on the knowledge information from the open domain KG. From Table~\ref{rec_ablation_table}, we can also note that the top-1 recommendation performances of KBRD, COLING20, KGSF, and KECRS can be significantly improved, by replacing DBpedia with TMDKG. This demonstrates that the high-quality information in TMDKG can help recommendation models better understand the user's intentions and thus make more accurate item recommendations. With the knowledge from TMDKG, the proposed KECRS model significantly outperforms all baseline methods, in terms of all the evaluation metrics. For example, based on TMDKG, KECRS outperforms KBRD, COLING20, and KGCF by 18.88\%, 20.96\%, and 30.05\%, in terms of Precision@1, respectively.

\subsection{Response Generation Performances}
\subsubsection{Automatic Evaluation} 
The automatic evaluation results of different methods in the response generation task are shown in Table \ref{response_eval_table}. 
We can note that Transformer performs better than REDIAL, which demonstrates that Transformer is powerful to understand and generate natural language. KBRD performs better than the basic Transformer model, because it adds a vocabulary bias to fuse knowledge from DBpedia into the generated responses. Among all the baseline models, KGSF generates the most diverse responses, by exploiting both DBpedia and ConceptNet. 
The potential reason is that KGSF employs two additional KG-based attention layers to make the generative model focus more on items and relevant entities in DBpedia and ConceptNet. Moreover, the proposed KECRS model outperforms all baseline methods with a large margin, in terms of all automatic evaluation metrics. 
This demonstrates that the proposed BOE loss, infusion loss, and newly constructed TMDKG can work jointly to generate more diverse and informative responses.

\subsubsection{Human Evaluation}
In this work, we also perform human evaluation to study the response generation performances of different methods. The human evaluation results are summarized in Table~\ref{response_eval_table}. 
We can make similar observations as in the automatic evaluation scenario, \ie Transformer and KBRD perform better than REDIAL, KGSF performs best among all baseline models, and KECRS performs better than KGSF with a large margin. Moreover, we can note that \textit{Fluency} is relevantly higher compared to \textit{Informativeness} and \textit{Relevancy} for all models. This indicates that responses generated by these models are fluent and can be understood by human beings. However, responses generated by baseline models are more likely to be some generic responses, which are also called "safe responses" (\eg I haven't seen that one). By including additional supervision signals, aligning embeddings of word and entities, and introducing the high-quality KG, our proposed KECRS model can alleviate this issue. Overall, KECRS can understand the dialogue context and generate fluent, relevant, and informative responses.

\begin{table}[]
    \centering
    \caption{Response generation performances of KGSF and different variants of KECRS. $^{*}$ indicates the improvement over the best baseline method is statistically significant with $p < 0.01$ using student \textit{t-test}.}
    \begin{tabular}{c|c|c|c|c}
    \hline
        Model & Dist-2 & Dist-3 & Dist-4 & AEN \\
    \hline 
        KGSF & 0.33 & 0.49 & 0.61 & 0.17\\
    \hline
        KECRS\textsubscript{w/o BOE} & 0.31 & 0.64 & 0.87 & 0.20\\
    \hline 
        KECRS\textsubscript{w/o Infuse} & 0.36 & 0.69 & 0.95 & 0.21\\
    \hline 
        KECRS & \textbf{0.48}$^*$ & \textbf{0.91}$^*$ & \textbf{1.23}$^*$ & \textbf{0.34}$^*$\\
    \hline     
    \end{tabular}
    \label{res_ablation_study}
\end{table}

\subsubsection{Ablation Study}
To better understand the effectiveness of each component in KECRS, we study the performances of the following two variants of KECRS:  1) \textbf{KECRS\textsubscript{w/o BOE}}: removing the BOE loss by setting $\lambda_1=0$; 2) \textbf{KECRS\textsubscript{w/o Infusion}}: removing the infusion loss by setting $\lambda_2=0$.

Table~\ref{res_ablation_study} summarizes the response generation performances of KGSF, KECRS\textsubscript{w/o BOE}, KECRS\textsubscript{w/o Infusion}, and KECRS, in terms of Distinct n-gram (n=2,3,4) and Average Entity Number (AEN). From Table~\ref{res_ablation_study}, we can note that KECRS outperforms KECRS\textsubscript{w/o BOE}. This indicates that the proposed BOE loss can help the model learn to generate responses not only from ground truth but also from the knowledge information in KG. This observation also demonstrates the lack of information about recommended items in the original dataset. Moreover, KECRS\textsubscript{w/o Infusion} performs poorer than KECRS. This indicates that bridging the representation space gap between the word embeddings and entity embeddings also helps improve the model performances. Compared with KGSF, both KECRS\textsubscript{w/o BOE} and KECRS\textsubscript{w/o Infusion} can achieve better performances in terms of most metrics, which again demonstrates the effectiveness of the proposed KECRS model.

\begin{figure}[]
    \begin{minipage}[t]{0.49\columnwidth}
        {\centering\includegraphics[width=0.95\linewidth]{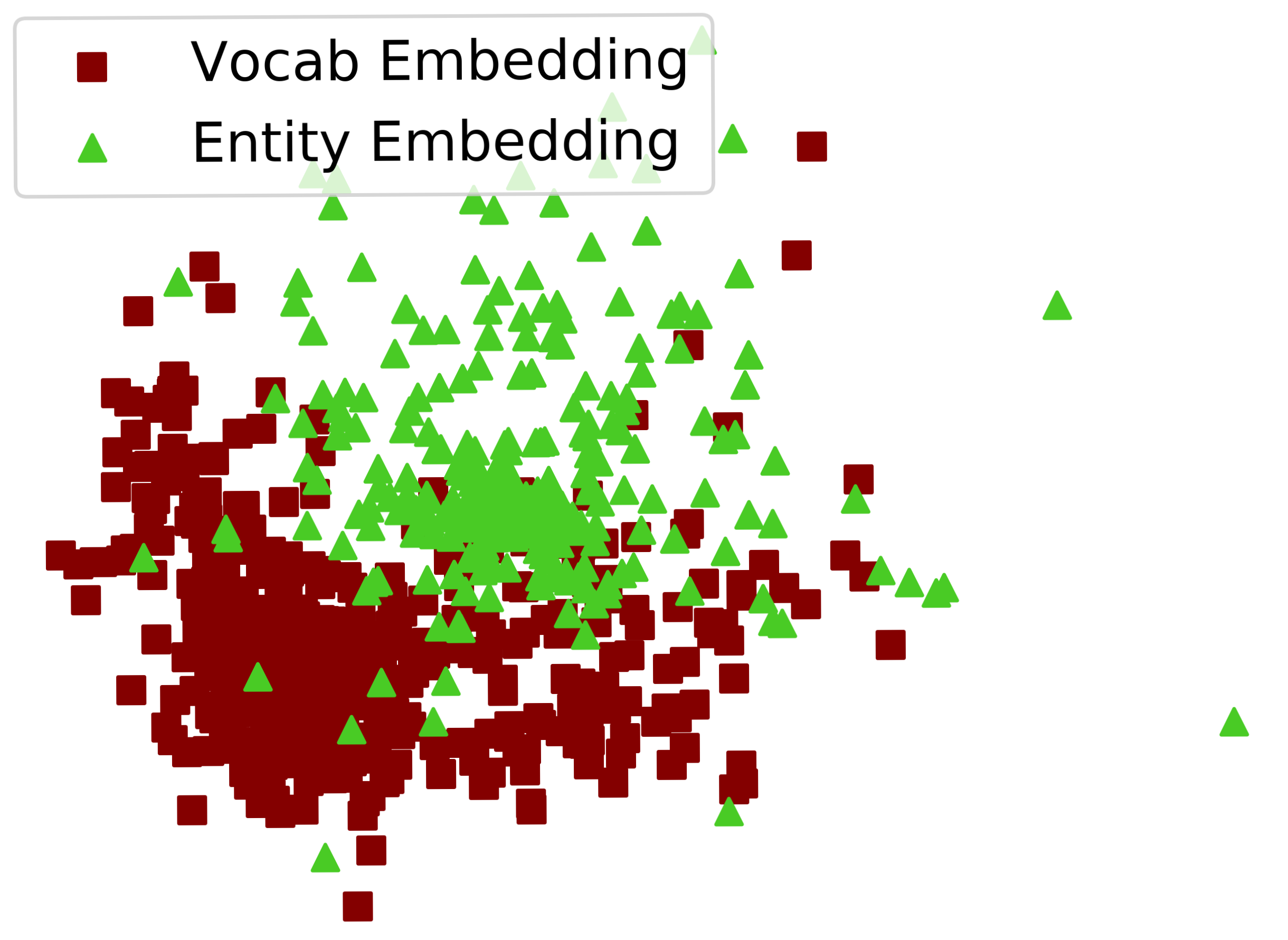}
            \label{KECRS w/o infuse}
        }
        \caption*{\raggedright(a) KECRS w/o infusion loss}
    \end{minipage}
    \begin{minipage}[t]{0.49\columnwidth}
        {\centering\includegraphics[width=0.95\linewidth]{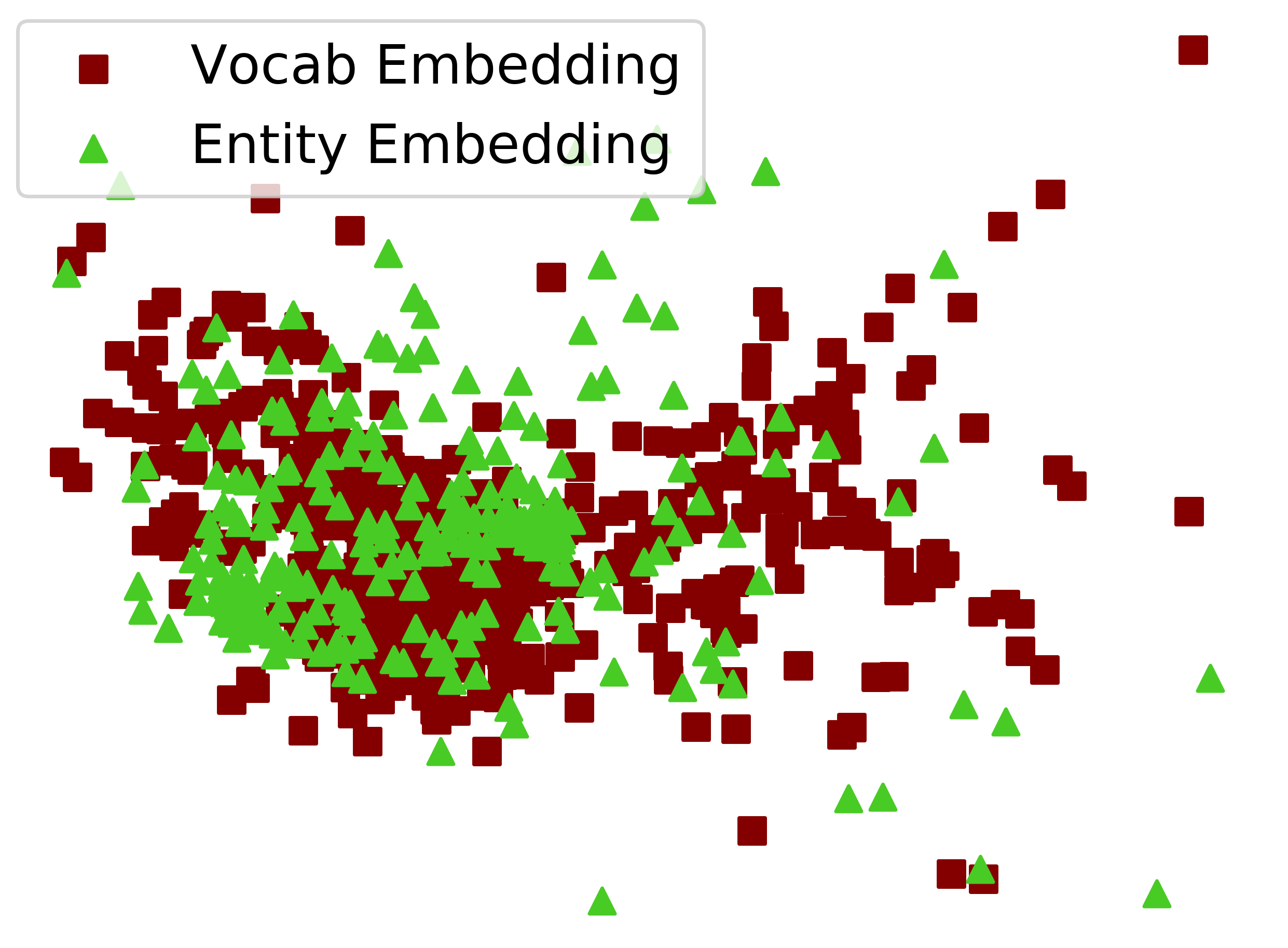}
            \label{KECRS w infuse}
        }
        \caption*{\raggedright(b) KECRS with infusion loss}
    \end{minipage}
    \caption{2D plot of word embeddings in response module and entity embeddings in recommendation module after PCA. Red points represent word embeddings and green points represent entity embeddings.}
    \label{fig:infuse_results}
\end{figure}

To further study the effect of the infusion loss, we draw the 2D plot of word embeddings in the response module and entity embeddings in the recommendation module after PCA in Figure~\ref{fig:infuse_results}. By using the infusion loss, two embeddings tend to cluster together, which indicates the infusion loss can bridge the gap between two representation spaces.



\subsubsection{Parameter Sensitivity Study}
\begin{figure}[]
    \centering
    \includegraphics[width=0.98\columnwidth]{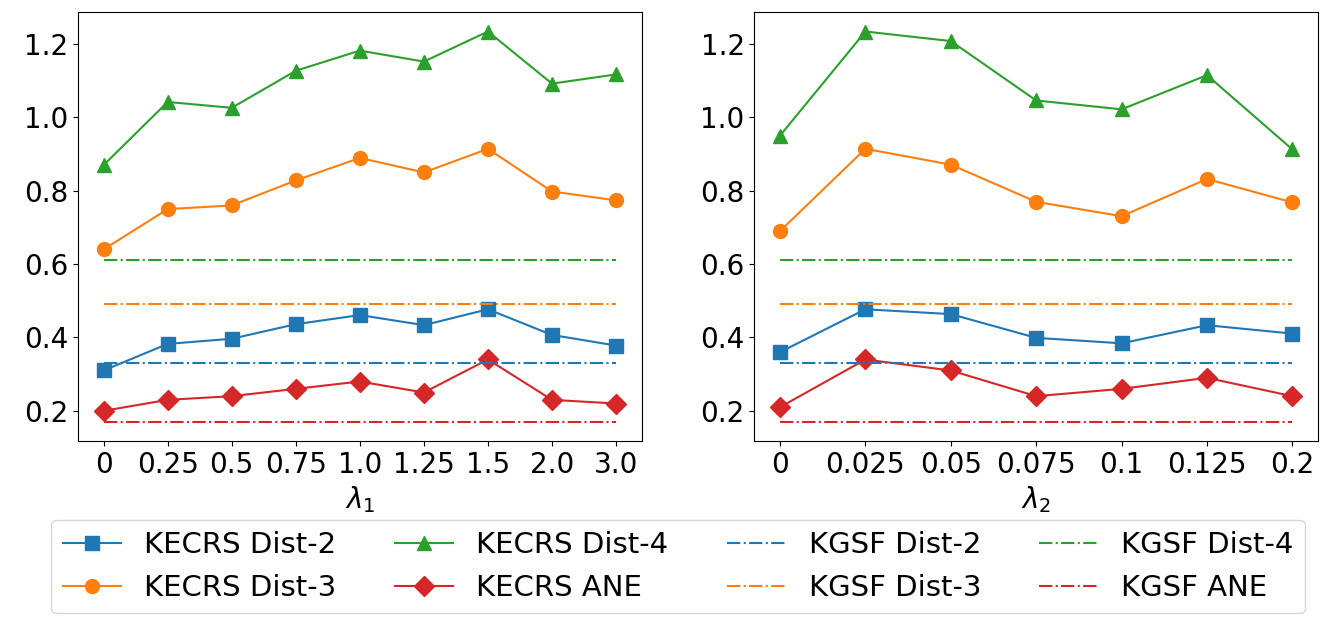}
    \caption{Response generation performance trends of KECRS with respect to different settings of $\lambda_1$ and $\lambda_2$.}
    \label{fig:lambda}
\end{figure}

$\lambda_1$ and $\lambda_2$ are two import hyper-parameters used to determine the weights of different losses when training the response generation module. We conduct experiments to study the impacts of these two hyper-parameters as shown in Figure~\ref{fig:lambda}. We can note that, with the increase of $\lambda_1$, the performances of KECRS improve first and start to decrease when $\lambda_1$ is larger than 1.5. As the primary objective of KECRS is learning from the ground truth responses instead of KG, too large $\lambda_1$ may lead to negative impacts on the performances of KECRS. For $\lambda_2$, when increasing it, the infusion loss may cause the word embedding over-smooth and decrease the response generation performances achieved by KECRS. $\lambda_3$ is a hyper-parameter only used in the testing phase. It is used to introduce position-irrelevant entities into the generated responses. As too large of $\lambda_3$ may hurt the fluency of generated responses, we empirically set it to 0.1.

\begin{table}[]
    \centering
    \caption{Response generation performances achieved by different methods with different knowledge graphs. $^{*}$ denotes that the improvement over the best baseline method is statistically significant with $p < 0.01$ using student \textit{t-test}.}
    \begin{tabular}{c|c|c|c|c|c}
    \hline
        Model & KG & Dist-2 & Dist-3 & Dist-4 & AEN \\
    \hline 
        \multirow{2}*{KBRD} & DBpedia& 0.26 & 0.30 & 0.45 & 0.15\\
         & TMDKG & 0.31 & 0.38 & 0.52 & 0.19\\
    \hline
        \multirow{2}*{KGSF}  & DBpedia &0.33 & 0.49 & 0.61 & 0.17\\
         & TMDKG & 0.38 & 0.61 & 0.73 & 0.20\\
    \hline 
        \multirow{2}*{KECRS} & DBpedia & 0.40  & 0.79 & 0.89 & 0.25\\
         & TMDKG & \textbf{0.48}$^*$ & \textbf{0.91}$^*$ & \textbf{1.23}$^*$ & \textbf{0.34}$^*$\\
    \hline     
    \end{tabular}
    \label{res_ablation_study_different_kg}
\end{table}

\subsubsection{Impacts of Knowledge Graph}
Similar to the experiments in Section~\ref{ss:recperformances}, we replace the DBpedia used in KBRD and KGSF by TMDKG, and also replace TMDKG used in the KECRS model by DBpedia. As shown in Table~\ref{res_ablation_study_different_kg}, the response generation performances of KBRD, KGSF, and KECRS, in terms of all evaluation metrics, can be improved by using TMDKG instead of DBpedia. This demonstrates that TMDKG contains more sufficient and less noisy information than DBpedia for the response generation task. This observation is also consistent with the observed patterns in the recommendation module (Section~\ref{ss:recperformances}). In addition, we can also note that KECRS consistently achieves better response generation performances than KBRD and KGSF in terms of all metrics, based on DBpedia and TMDKG. This demonstrates the robustness of the proposed KECRS model in the response generation task.


\subsection{Case Study}
Table~\ref{case_study} shows a case study about the conversations generated by different models.
Compared with baseline models, KECRS can not only capture user's intentions to provide the corresponding recommendations, but also produce informative responses regarding the recommended item.
As shown in Table ~\ref{case_study}, when the user expresses her preferences on "\textit{A Nightmare on Elm Street (1984)}" and "\textit{The Last House on the Left (1972)}", KECRS can understand the user may like thrillers. Thus, KECRS recommends another thriller "\textit{It (2017)}" to her. While providing the item recommendation, KECRS also makes the conversation more informative and natural by including \textit{"if you like horror movies with murder mystery"} in the response.

\begin{table}[]
    \caption{Case Study. The user shows her preferences on two thrillers. KECRS captures this information and recommends another thriller with relevant information.}
    \begin{tabular}{l p{5.5cm}}
    \hline
        Seeker: & Hi \\
        Recommender: & Hi! What kind of movies do you like?\\
        Seeker: & I am looking for a movie recommendation. When I was younger, I really enjoyed the \textit{\textbf{A Nightmare on Elm Street (1984)}}. I also enjoyed watching \textit{\textbf{The Last House on the Left (1972)}}.\\
    \hline
        Transformer: & I haven't seen that one. \\
    \hline 
        KBRD: & I haven't seen that one yet.\\
    \hline 
        KGSF: & I don't know that one. Have you seen \textit{\textbf{Boo! A Madea Halloween (2016)}}?\\
    \hline 
        KECRS (Ours): & You'll have to check out \textit{\textbf{It (2017)}} if you like \textit{\textbf{horror}} movies with \textit{\textbf{murder mystery}}.\\
    \hline 
        Human: & Oh, you like scary movies? I recently watched Happy Death Day  (2017). It was really good for a new “scary movie”.\\
    \hline
    \end{tabular}
    \label{case_study}
\end{table}

\section{Conclusion and Future Work}
In this paper, we first reformulate the conversational recommendation task. Then, we propose a novel Knowledge-Enriched Conversational Recommendation System (KECRS). Specifically, we develop the Bag-of-Entity (BOE) loss and the infusion loss to improve the response generation performances of the proposed KECRS model. Moreover, we facilitate our study by constructing a high-quality KG, namely TMDKG. The experimental results demonstrate that the proposed BOE loss can guide the model to generate more knowledge-enriched responses by selecting entities in KG, and the infusion loss can align the vocabulary and entity embeddings into the same space. In addition, the newly built TMDKG also helps both KECRS and baseline methods better understand the user's preferences and generate more informative responses. Overall, KECRS usually achieves superior performances on both recommendation accuracy and response quality than state-of-the-art baselines. For future work, we would like to investigate how to use keywords to conduct the conversation from chit-chat to the recommendation~\cite{zhou2020towards,liu2020towards}. Moreover, we are also interested in using external knowledge (\eg KG) to modify the REDIAL dataset and make responses more relevant to the recommended items \cite{jannach2020}. 




\bibliographystyle{ACM-Reference-Format}
\bibliography{refer_short}


\begin{thebibliography}{40}


\ifx \showCODEN    \undefined \def \showCODEN     #1{\unskip}     \fi
\ifx \showDOI      \undefined \def \showDOI       #1{#1}\fi
\ifx \showISBNx    \undefined \def \showISBNx     #1{\unskip}     \fi
\ifx \showISBNxiii \undefined \def \showISBNxiii  #1{\unskip}     \fi
\ifx \showISSN     \undefined \def \showISSN      #1{\unskip}     \fi
\ifx \showLCCN     \undefined \def \showLCCN      #1{\unskip}     \fi
\ifx \shownote     \undefined \def \shownote      #1{#1}          \fi
\ifx \showarticletitle \undefined \def \showarticletitle #1{#1}   \fi
\ifx \showURL      \undefined \def \showURL       {\relax}        \fi
\providecommand\bibfield[2]{#2}
\providecommand\bibinfo[2]{#2}
\providecommand\natexlab[1]{#1}
\providecommand\showeprint[2][]{arXiv:#2}

\bibitem[\protect\citeauthoryear{Bobadilla, Ortega, Hernando, and
  Guti{\'e}rrez}{Bobadilla et~al\mbox{.}}{2013}]%
        {bobadilla2013recommender}
\bibfield{author}{\bibinfo{person}{Jes{\'u}s Bobadilla},
  \bibinfo{person}{Fernando Ortega}, \bibinfo{person}{Antonio Hernando}, {and}
  \bibinfo{person}{Abraham Guti{\'e}rrez}.} \bibinfo{year}{2013}\natexlab{}.
\newblock \showarticletitle{Recommender systems survey}.
\newblock \bibinfo{journal}{\emph{Knowledge-Based Systems}}
  \bibinfo{volume}{46} (\bibinfo{year}{2013}), \bibinfo{pages}{109--132}.
\newblock


\bibitem[\protect\citeauthoryear{Chen, Lin, Zhang, Ding, Cen, Yang, and
  Tang}{Chen et~al\mbox{.}}{2019}]%
        {chen2019towards}
\bibfield{author}{\bibinfo{person}{Qibin Chen}, \bibinfo{person}{Junyang Lin},
  \bibinfo{person}{Yichang Zhang}, \bibinfo{person}{Ming Ding},
  \bibinfo{person}{Yukuo Cen}, \bibinfo{person}{Hongxia Yang}, {and}
  \bibinfo{person}{Jie Tang}.} \bibinfo{year}{2019}\natexlab{}.
\newblock \showarticletitle{{Towards knowledge-based recommender dialog
  system}}.
\newblock \bibinfo{journal}{\emph{arXiv preprint arXiv:1908.05391}}
  (\bibinfo{year}{2019}).
\newblock


\bibitem[\protect\citeauthoryear{Christakopoulou, Beutel, Li, Jain, and
  Chi}{Christakopoulou et~al\mbox{.}}{2018}]%
        {christakopoulou2018q}
\bibfield{author}{\bibinfo{person}{Konstantina Christakopoulou},
  \bibinfo{person}{Alex Beutel}, \bibinfo{person}{Rui Li},
  \bibinfo{person}{Sagar Jain}, {and} \bibinfo{person}{Ed~H. Chi}.}
  \bibinfo{year}{2018}\natexlab{}.
\newblock \showarticletitle{{Q\&R: A two-stage approach toward interactive
  recommendation}}. In \bibinfo{booktitle}{\emph{SIGKDD}}.
  \bibinfo{pages}{139--148}.
\newblock


\bibitem[\protect\citeauthoryear{Christakopoulou, Radlinski, and
  Hofmann}{Christakopoulou et~al\mbox{.}}{2016}]%
        {christakopoulou2016towards}
\bibfield{author}{\bibinfo{person}{Konstantina Christakopoulou},
  \bibinfo{person}{Filip Radlinski}, {and} \bibinfo{person}{Katja Hofmann}.}
  \bibinfo{year}{2016}\natexlab{}.
\newblock \showarticletitle{Towards conversational recommender systems}. In
  \bibinfo{booktitle}{\emph{SIGKDD}}. \bibinfo{pages}{815--824}.
\newblock


\bibitem[\protect\citeauthoryear{Devlin, Chang, Lee, and Toutanova}{Devlin
  et~al\mbox{.}}{2018}]%
        {devlin2018bert}
\bibfield{author}{\bibinfo{person}{Jacob Devlin}, \bibinfo{person}{Ming-Wei
  Chang}, \bibinfo{person}{Kenton Lee}, {and} \bibinfo{person}{Kristina
  Toutanova}.} \bibinfo{year}{2018}\natexlab{}.
\newblock \showarticletitle{{Bert: Pre-training of deep bidirectional
  Transformers for language understanding}}.
\newblock \bibinfo{journal}{\emph{arXiv preprint arXiv:1810.04805}}
  (\bibinfo{year}{2018}).
\newblock


\bibitem[\protect\citeauthoryear{Dinan, Roller, Shuster, Fan, Auli, and
  Weston}{Dinan et~al\mbox{.}}{2018}]%
        {dinan2018wizard}
\bibfield{author}{\bibinfo{person}{Emily Dinan}, \bibinfo{person}{Stephen
  Roller}, \bibinfo{person}{Kurt Shuster}, \bibinfo{person}{Angela Fan},
  \bibinfo{person}{Michael Auli}, {and} \bibinfo{person}{Jason Weston}.}
  \bibinfo{year}{2018}\natexlab{}.
\newblock \showarticletitle{{Wizard of wikipedia: Knowledge-powered
  conversational agents}}.
\newblock \bibinfo{journal}{\emph{arXiv preprint arXiv:1811.01241}}
  (\bibinfo{year}{2018}).
\newblock


\bibitem[\protect\citeauthoryear{Gao, Lei, He, de~Rijke, and Chua}{Gao
  et~al\mbox{.}}{2021}]%
        {gao2021advances}
\bibfield{author}{\bibinfo{person}{Chongming Gao}, \bibinfo{person}{Wenqiang
  Lei}, \bibinfo{person}{Xiangnan He}, \bibinfo{person}{Maarten de Rijke},
  {and} \bibinfo{person}{Tat-Seng Chua}.} \bibinfo{year}{2021}\natexlab{}.
\newblock \showarticletitle{{Advances and challenges in conversational
  recommender systems: A survey}}.
\newblock \bibinfo{journal}{\emph{{arXiv preprint arXiv:2101.09459}}}
  (\bibinfo{year}{2021}).
\newblock


\bibitem[\protect\citeauthoryear{Hayati, Kang, Zhu, Shi, and Yu}{Hayati
  et~al\mbox{.}}{2020}]%
        {hayati2020inspired}
\bibfield{author}{\bibinfo{person}{Shirley~Anugrah Hayati},
  \bibinfo{person}{Dongyeop Kang}, \bibinfo{person}{Qingxiaoyang Zhu},
  \bibinfo{person}{Weiyan Shi}, {and} \bibinfo{person}{Zhou Yu}.}
  \bibinfo{year}{2020}\natexlab{}.
\newblock \showarticletitle{{INSPIRED: Toward sociable recommendation dialog
  systems}}.
\newblock \bibinfo{journal}{\emph{{arXiv preprint arXiv:2009.14306}}}
  (\bibinfo{year}{2020}).
\newblock


\bibitem[\protect\citeauthoryear{Jannach and Manzoor}{Jannach and
  Manzoor}{2020}]%
        {jannach2020}
\bibfield{author}{\bibinfo{person}{Dietmar Jannach} {and}
  \bibinfo{person}{Ahtsham Manzoor}.} \bibinfo{year}{2020}\natexlab{}.
\newblock \showarticletitle{{End-to-End learning for conversational
  recommendation: A long way to go?}}. In \bibinfo{booktitle}{\emph{RecSys}}.
\newblock


\bibitem[\protect\citeauthoryear{Jannach, Manzoor, Cai, and Chen}{Jannach
  et~al\mbox{.}}{2020}]%
        {jannach2020survey}
\bibfield{author}{\bibinfo{person}{Dietmar Jannach}, \bibinfo{person}{Ahtsham
  Manzoor}, \bibinfo{person}{Wanling Cai}, {and} \bibinfo{person}{Li Chen}.}
  \bibinfo{year}{2020}\natexlab{}.
\newblock \showarticletitle{A survey on conversational recommender systems}.
\newblock \bibinfo{journal}{\emph{arXiv preprint arXiv:2004.00646}}
  (\bibinfo{year}{2020}).
\newblock


\bibitem[\protect\citeauthoryear{Ji, Pi, Wei, Xiong, Wo{\'z}niak, and
  Damasevicius}{Ji et~al\mbox{.}}{2019}]%
        {ji2019recommendation}
\bibfield{author}{\bibinfo{person}{Zhenyan Ji}, \bibinfo{person}{Huaiyu Pi},
  \bibinfo{person}{Wei Wei}, \bibinfo{person}{Bo Xiong},
  \bibinfo{person}{Marcin Wo{\'z}niak}, {and} \bibinfo{person}{Robertas
  Damasevicius}.} \bibinfo{year}{2019}\natexlab{}.
\newblock \showarticletitle{{Recommendation based on review texts and social
  communities: A hybrid model}}.
\newblock \bibinfo{journal}{\emph{{IEEE Access}}}  \bibinfo{volume}{7}
  (\bibinfo{year}{2019}), \bibinfo{pages}{40416--40427}.
\newblock


\bibitem[\protect\citeauthoryear{Kingma and Ba}{Kingma and Ba}{2014}]%
        {kingma2014adam}
\bibfield{author}{\bibinfo{person}{Diederik~P. Kingma} {and}
  \bibinfo{person}{Jimmy Ba}.} \bibinfo{year}{2014}\natexlab{}.
\newblock \showarticletitle{{Adam: A method for stochastic optimization}}.
\newblock \bibinfo{journal}{\emph{{arXiv preprint arXiv:1412.6980}}}
  (\bibinfo{year}{2014}).
\newblock


\bibitem[\protect\citeauthoryear{Koncel-Kedziorski, Bekal, Luan, Lapata, and
  Hajishirzi}{Koncel-Kedziorski et~al\mbox{.}}{2019}]%
        {koncel2019text}
\bibfield{author}{\bibinfo{person}{Rik Koncel-Kedziorski},
  \bibinfo{person}{Dhanush Bekal}, \bibinfo{person}{Yi Luan},
  \bibinfo{person}{Mirella Lapata}, {and} \bibinfo{person}{Hannaneh
  Hajishirzi}.} \bibinfo{year}{2019}\natexlab{}.
\newblock \showarticletitle{Text generation from knowledge graphs with graph
  transformers}.
\newblock \bibinfo{journal}{\emph{arXiv preprint arXiv:1904.02342}}
  (\bibinfo{year}{2019}).
\newblock


\bibitem[\protect\citeauthoryear{Lehmann, Isele, Jakob, Jentzsch, Kontokostas,
  Mendes, and et~al.}{Lehmann et~al\mbox{.}}{2014}]%
        {dbpedia}
\bibfield{author}{\bibinfo{person}{Jens Lehmann}, \bibinfo{person}{Robert
  Isele}, \bibinfo{person}{Max Jakob}, \bibinfo{person}{Anja Jentzsch},
  \bibinfo{person}{Dimitris Kontokostas}, \bibinfo{person}{Pablo Mendes}, {and}
  \bibinfo{person}{et al.}} \bibinfo{year}{2014}\natexlab{}.
\newblock \showarticletitle{{DBpedia - A large-scale, multilingual knowledge
  base extracted from Wikipedia}}.
\newblock \bibinfo{journal}{\emph{Semantic Web Journal}}  \bibinfo{volume}{6}
  (\bibinfo{year}{2014}).
\newblock


\bibitem[\protect\citeauthoryear{Lei, He, Miao, Wu, Hong, Kan, and Chua}{Lei
  et~al\mbox{.}}{2020a}]%
        {lei2020estimation}
\bibfield{author}{\bibinfo{person}{Wenqiang Lei}, \bibinfo{person}{Xiangnan
  He}, \bibinfo{person}{Yisong Miao}, \bibinfo{person}{Qingyun Wu},
  \bibinfo{person}{Richang Hong}, \bibinfo{person}{Min-Yen Kan}, {and}
  \bibinfo{person}{Tat-Seng Chua}.} \bibinfo{year}{2020}\natexlab{a}.
\newblock \showarticletitle{{Estimation-action-reflection: Towards deep
  interaction between conversational and recommender systems}}. In
  \bibinfo{booktitle}{\emph{WSDM}}. \bibinfo{pages}{304--312}.
\newblock


\bibitem[\protect\citeauthoryear{Lei, Zhang, He, Miao, Wang, Chen, and
  Chua}{Lei et~al\mbox{.}}{2020b}]%
        {lei2020interactive}
\bibfield{author}{\bibinfo{person}{Wenqiang Lei}, \bibinfo{person}{Gangyi
  Zhang}, \bibinfo{person}{Xiangnan He}, \bibinfo{person}{Yisong Miao},
  \bibinfo{person}{Xiang Wang}, \bibinfo{person}{Liang Chen}, {and}
  \bibinfo{person}{Tat-Seng Chua}.} \bibinfo{year}{2020}\natexlab{b}.
\newblock \showarticletitle{{Interactive path reasoning on graph for
  conversational recommendation}}. In \bibinfo{booktitle}{\emph{SIGKDD}}.
  \bibinfo{pages}{2073--2083}.
\newblock


\bibitem[\protect\citeauthoryear{Li, Kahou, Schulz, Michalski, Charlin, and
  Pal}{Li et~al\mbox{.}}{2018}]%
        {li2018towards}
\bibfield{author}{\bibinfo{person}{Raymond Li},
  \bibinfo{person}{Samira~Ebrahimi Kahou}, \bibinfo{person}{Hannes Schulz},
  \bibinfo{person}{Vincent Michalski}, \bibinfo{person}{Laurent Charlin}, {and}
  \bibinfo{person}{Chris Pal}.} \bibinfo{year}{2018}\natexlab{}.
\newblock \showarticletitle{{Towards deep conversational recommendations}}. In
  \bibinfo{booktitle}{\emph{NeurIPS}}. \bibinfo{pages}{9725--9735}.
\newblock


\bibitem[\protect\citeauthoryear{Liu, Wang, Xu, Peng, and Jiao}{Liu
  et~al\mbox{.}}{2020c}]%
        {liu2020neural}
\bibfield{author}{\bibinfo{person}{Hongtao Liu}, \bibinfo{person}{Wenjun Wang},
  \bibinfo{person}{Hongyan Xu}, \bibinfo{person}{Qiyao Peng}, {and}
  \bibinfo{person}{Pengfei Jiao}.} \bibinfo{year}{2020}\natexlab{c}.
\newblock \showarticletitle{Neural unified review recommendation with cross
  attention}. In \bibinfo{booktitle}{\emph{SIGIR}}.
  \bibinfo{pages}{1789--1792}.
\newblock


\bibitem[\protect\citeauthoryear{Liu, Ounis, Macdonald, and Meng}{Liu
  et~al\mbox{.}}{2020a}]%
        {liu2020heterogeneous}
\bibfield{author}{\bibinfo{person}{Siwei Liu}, \bibinfo{person}{Iadh Ounis},
  \bibinfo{person}{Craig Macdonald}, {and} \bibinfo{person}{Zaiqiao Meng}.}
  \bibinfo{year}{2020}\natexlab{a}.
\newblock \showarticletitle{A heterogeneous graph neural model for cold-start
  recommendation}. In \bibinfo{booktitle}{\emph{SIGIR}}.
  \bibinfo{pages}{2029--2032}.
\newblock


\bibitem[\protect\citeauthoryear{Liu, Wang, Niu, Wu, Che, and Liu}{Liu
  et~al\mbox{.}}{2020b}]%
        {liu2020towards}
\bibfield{author}{\bibinfo{person}{Zeming Liu}, \bibinfo{person}{Haifeng Wang},
  \bibinfo{person}{Zheng-Yu Niu}, \bibinfo{person}{Hua Wu},
  \bibinfo{person}{Wanxiang Che}, {and} \bibinfo{person}{Ting Liu}.}
  \bibinfo{year}{2020}\natexlab{b}.
\newblock \showarticletitle{Towards conversational recommendation over
  multi-type dialogs}.
\newblock \bibinfo{journal}{\emph{arXiv preprint arXiv:2005.03954}}
  (\bibinfo{year}{2020}).
\newblock


\bibitem[\protect\citeauthoryear{Ma, Sun, Wang, and Lin}{Ma
  et~al\mbox{.}}{2018}]%
        {ma2018bag}
\bibfield{author}{\bibinfo{person}{Shuming Ma}, \bibinfo{person}{Xu Sun},
  \bibinfo{person}{Yizhong Wang}, {and} \bibinfo{person}{Junyang Lin}.}
  \bibinfo{year}{2018}\natexlab{}.
\newblock \showarticletitle{Bag-of-words as target for neural machine
  translation}.
\newblock \bibinfo{journal}{\emph{{arXiv preprint arXiv:1805.04871}}}
  (\bibinfo{year}{2018}).
\newblock


\bibitem[\protect\citeauthoryear{Madotto, Wu, and Fung}{Madotto
  et~al\mbox{.}}{2018}]%
        {madotto2018mem2seq}
\bibfield{author}{\bibinfo{person}{Andrea Madotto},
  \bibinfo{person}{Chien-Sheng Wu}, {and} \bibinfo{person}{Pascale Fung}.}
  \bibinfo{year}{2018}\natexlab{}.
\newblock \showarticletitle{{Mem2seq: Effectively incorporating knowledge bases
  into end-to-end task-oriented dialog systems}}.
\newblock \bibinfo{journal}{\emph{arXiv preprint arXiv:1804.08217}}
  (\bibinfo{year}{2018}).
\newblock


\bibitem[\protect\citeauthoryear{Paszke, Gross, Massa, Lerer, Bradbury, Chanan,
  and et~al.}{Paszke et~al\mbox{.}}{2019}]%
        {pytorch19}
\bibfield{author}{\bibinfo{person}{Adam Paszke}, \bibinfo{person}{Sam Gross},
  \bibinfo{person}{Francisco Massa}, \bibinfo{person}{Adam Lerer},
  \bibinfo{person}{James Bradbury}, \bibinfo{person}{Gregory Chanan}, {and}
  \bibinfo{person}{et al.}} \bibinfo{year}{2019}\natexlab{}.
\newblock \showarticletitle{{PyTorch: An Imperative Style, High-Performance
  Deep Learning Library}}. In \bibinfo{booktitle}{\emph{NeurIPS}}.
  \bibinfo{pages}{8024--8035}.
\newblock


\bibitem[\protect\citeauthoryear{Ren, Yin, Chen, Wang, Hung, Huang, and
  Zhang}{Ren et~al\mbox{.}}{2020}]%
        {ren2020crsal}
\bibfield{author}{\bibinfo{person}{Xuhui Ren}, \bibinfo{person}{Hongzhi Yin},
  \bibinfo{person}{Tong Chen}, \bibinfo{person}{Hao Wang},
  \bibinfo{person}{Nguyen Quoc~Viet Hung}, \bibinfo{person}{Zi Huang}, {and}
  \bibinfo{person}{Xiangliang Zhang}.} \bibinfo{year}{2020}\natexlab{}.
\newblock \showarticletitle{{Crsal: Conversational recommender systems with
  adversarial learning}}.
\newblock \bibinfo{journal}{\emph{ACM Transactions on Information Systems
  (TOIS)}} \bibinfo{volume}{38}, \bibinfo{number}{4} (\bibinfo{year}{2020}),
  \bibinfo{pages}{1--40}.
\newblock


\bibitem[\protect\citeauthoryear{Sarkar, Goswami, Arcan, and McCrae}{Sarkar
  et~al\mbox{.}}{2020}]%
        {sarkar2020suggest}
\bibfield{author}{\bibinfo{person}{Rajdeep Sarkar}, \bibinfo{person}{Koustava
  Goswami}, \bibinfo{person}{Mihael Arcan}, {and} \bibinfo{person}{John~Philip
  McCrae}.} \bibinfo{year}{2020}\natexlab{}.
\newblock \showarticletitle{{Suggest me a movie for tonight: Leveraging
  knowledge graphs for conversational recommendation}}. In
  \bibinfo{booktitle}{\emph{COLING}}. \bibinfo{pages}{4179--4189}.
\newblock


\bibitem[\protect\citeauthoryear{Schafer, Konstan, and Riedl}{Schafer
  et~al\mbox{.}}{2001}]%
        {schafer2001commerce}
\bibfield{author}{\bibinfo{person}{J.~Ben Schafer}, \bibinfo{person}{Joseph~A.
  Konstan}, {and} \bibinfo{person}{John Riedl}.}
  \bibinfo{year}{2001}\natexlab{}.
\newblock \showarticletitle{E-commerce recommendation applications}.
\newblock \bibinfo{journal}{\emph{Data mining and knowledge discovery}}
  \bibinfo{volume}{5}, \bibinfo{number}{1-2} (\bibinfo{year}{2001}),
  \bibinfo{pages}{115--153}.
\newblock


\bibitem[\protect\citeauthoryear{Schlichtkrull, Kipf, Bloem, Van Den~Berg,
  Titov, and Welling}{Schlichtkrull et~al\mbox{.}}{2017}]%
        {schlichtkrull2018modeling}
\bibfield{author}{\bibinfo{person}{Michael Schlichtkrull},
  \bibinfo{person}{Thomas~N. Kipf}, \bibinfo{person}{Peter Bloem},
  \bibinfo{person}{Rianne Van Den~Berg}, \bibinfo{person}{Ivan Titov}, {and}
  \bibinfo{person}{Max Welling}.} \bibinfo{year}{2017}\natexlab{}.
\newblock \showarticletitle{Modeling relational data with graph convolutional
  networks}.
\newblock  (\bibinfo{year}{2017}).
\newblock


\bibitem[\protect\citeauthoryear{Serban, Sordoni, Lowe, Charlin, Pineau,
  Courville, and Bengio}{Serban et~al\mbox{.}}{2016}]%
        {serban2016hierarchical}
\bibfield{author}{\bibinfo{person}{Iulian~Vlad Serban},
  \bibinfo{person}{Alessandro Sordoni}, \bibinfo{person}{Ryan Lowe},
  \bibinfo{person}{Laurent Charlin}, \bibinfo{person}{Joelle Pineau},
  \bibinfo{person}{Aaron Courville}, {and} \bibinfo{person}{Yoshua Bengio}.}
  \bibinfo{year}{2016}\natexlab{}.
\newblock \showarticletitle{A hierarchical latent variable encoder-decoder
  model for generating dialogues}.
\newblock \bibinfo{journal}{\emph{{arXiv preprint arXiv:1605.06069}}}
  (\bibinfo{year}{2016}).
\newblock


\bibitem[\protect\citeauthoryear{Sun and Zhang}{Sun and Zhang}{2018}]%
        {sun2018conversational}
\bibfield{author}{\bibinfo{person}{Yueming Sun} {and} \bibinfo{person}{Yi
  Zhang}.} \bibinfo{year}{2018}\natexlab{}.
\newblock \showarticletitle{Conversational recommender system}. In
  \bibinfo{booktitle}{\emph{SIGIR}}. \bibinfo{pages}{235--244}.
\newblock


\bibitem[\protect\citeauthoryear{Vaswani, Shazeer, Parmar, Uszkoreit, Jones,
  Gomez, Kaiser, and Polosukhin}{Vaswani et~al\mbox{.}}{2017}]%
        {vaswani2017attention}
\bibfield{author}{\bibinfo{person}{Ashish Vaswani}, \bibinfo{person}{Noam
  Shazeer}, \bibinfo{person}{Niki Parmar}, \bibinfo{person}{Jakob Uszkoreit},
  \bibinfo{person}{Llion Jones}, \bibinfo{person}{Aidan~N. Gomez},
  \bibinfo{person}{Lukasz Kaiser}, {and} \bibinfo{person}{Illia Polosukhin}.}
  \bibinfo{year}{2017}\natexlab{}.
\newblock \showarticletitle{{Attention is all you need}}.
\newblock \bibinfo{journal}{\emph{arXiv preprint arXiv:1706.03762}}
  (\bibinfo{year}{2017}).
\newblock


\bibitem[\protect\citeauthoryear{Wu, Socher, and Xiong}{Wu
  et~al\mbox{.}}{2019}]%
        {wu2019global}
\bibfield{author}{\bibinfo{person}{Chien-Sheng Wu}, \bibinfo{person}{Richard
  Socher}, {and} \bibinfo{person}{Caiming Xiong}.}
  \bibinfo{year}{2019}\natexlab{}.
\newblock \showarticletitle{Global-to-local memory pointer networks for
  task-oriented dialogue}.
\newblock \bibinfo{journal}{\emph{{arXiv preprint arXiv:1901.04713}}}
  (\bibinfo{year}{2019}).
\newblock


\bibitem[\protect\citeauthoryear{Xu, Li, Tian, Sonobe, Kawarabayashi, and
  Jegelka}{Xu et~al\mbox{.}}{2018}]%
        {xu2018representation}
\bibfield{author}{\bibinfo{person}{Keyulu Xu}, \bibinfo{person}{Chengtao Li},
  \bibinfo{person}{Yonglong Tian}, \bibinfo{person}{Tomohiro Sonobe},
  \bibinfo{person}{Ken-ichi Kawarabayashi}, {and} \bibinfo{person}{Stefanie
  Jegelka}.} \bibinfo{year}{2018}\natexlab{}.
\newblock \showarticletitle{Representation learning on graphs with jumping
  knowledge networks}.
\newblock \bibinfo{journal}{\emph{arXiv preprint arXiv:1806.03536}}
  (\bibinfo{year}{2018}).
\newblock


\bibitem[\protect\citeauthoryear{Yan, Duan, Chen, Zhou, Zhou, and Li}{Yan
  et~al\mbox{.}}{2017}]%
        {yan2017building}
\bibfield{author}{\bibinfo{person}{Zhao Yan}, \bibinfo{person}{Nan Duan},
  \bibinfo{person}{Peng Chen}, \bibinfo{person}{Ming Zhou},
  \bibinfo{person}{Jianshe Zhou}, {and} \bibinfo{person}{Zhoujun Li}.}
  \bibinfo{year}{2017}\natexlab{}.
\newblock \showarticletitle{Building task-oriented dialogue systems for online
  shopping}. In \bibinfo{booktitle}{\emph{AAAI}}.
\newblock


\bibitem[\protect\citeauthoryear{Zhang, Yuan, Lian, Xie, and Ma}{Zhang
  et~al\mbox{.}}{2016}]%
        {zhang2016collaborative}
\bibfield{author}{\bibinfo{person}{Fuzheng Zhang},
  \bibinfo{person}{Nicholas~Jing Yuan}, \bibinfo{person}{Defu Lian},
  \bibinfo{person}{Xing Xie}, {and} \bibinfo{person}{Wei-Ying Ma}.}
  \bibinfo{year}{2016}\natexlab{}.
\newblock \showarticletitle{Collaborative knowledge base embedding for
  recommender systems}. In \bibinfo{booktitle}{\emph{SIGKDD}}.
  \bibinfo{pages}{353--362}.
\newblock


\bibitem[\protect\citeauthoryear{Zhang, Chen, Ai, Yang, and Croft}{Zhang
  et~al\mbox{.}}{2018}]%
        {zhang2018towards}
\bibfield{author}{\bibinfo{person}{Yongfeng Zhang}, \bibinfo{person}{Xu Chen},
  \bibinfo{person}{Qingyao Ai}, \bibinfo{person}{Liu Yang}, {and}
  \bibinfo{person}{W.~Bruce Croft}.} \bibinfo{year}{2018}\natexlab{}.
\newblock \showarticletitle{{Towards conversational search and recommendation:
  System ask, user respond}}. In \bibinfo{booktitle}{\emph{CIKM}}.
  \bibinfo{pages}{177--186}.
\newblock


\bibitem[\protect\citeauthoryear{Zhao, Zhao, and Eskenazi}{Zhao
  et~al\mbox{.}}{2017}]%
        {zhao2017learning}
\bibfield{author}{\bibinfo{person}{Tiancheng Zhao}, \bibinfo{person}{Ran Zhao},
  {and} \bibinfo{person}{Maxine Eskenazi}.} \bibinfo{year}{2017}\natexlab{}.
\newblock \showarticletitle{Learning discourse-level diversity for neural
  dialog models using conditional variational autoencoders}.
\newblock \bibinfo{journal}{\emph{arXiv preprint arXiv:1703.10960}}
  (\bibinfo{year}{2017}).
\newblock


\bibitem[\protect\citeauthoryear{Zhong, Wang, and Miao}{Zhong
  et~al\mbox{.}}{2019}]%
        {zhong2019affect}
\bibfield{author}{\bibinfo{person}{Peixiang Zhong}, \bibinfo{person}{Di Wang},
  {and} \bibinfo{person}{Chunyan Miao}.} \bibinfo{year}{2019}\natexlab{}.
\newblock \showarticletitle{An affect-rich neural conversational model with
  biased attention and weighted cross-entropy loss}. In
  \bibinfo{booktitle}{\emph{AAAI}}, Vol.~\bibinfo{volume}{33}.
  \bibinfo{pages}{7492--7500}.
\newblock


\bibitem[\protect\citeauthoryear{Zhou, Zhao, Bian, Zhou, Wen, and Yu}{Zhou
  et~al\mbox{.}}{2020a}]%
        {zhou2020improving}
\bibfield{author}{\bibinfo{person}{Kun Zhou}, \bibinfo{person}{Wayne~Xin Zhao},
  \bibinfo{person}{Shuqing Bian}, \bibinfo{person}{Yuanhang Zhou},
  \bibinfo{person}{Ji-Rong Wen}, {and} \bibinfo{person}{Jingsong Yu}.}
  \bibinfo{year}{2020}\natexlab{a}.
\newblock \showarticletitle{{Improving conversational recommender systems via
  knowledge graph based semantic fusion}}. In
  \bibinfo{booktitle}{\emph{SIGKDD}}. \bibinfo{pages}{1006--1014}.
\newblock


\bibitem[\protect\citeauthoryear{Zhou, Zhou, Zhao, Wang, and Wen}{Zhou
  et~al\mbox{.}}{2020b}]%
        {zhou2020towards}
\bibfield{author}{\bibinfo{person}{Kun Zhou}, \bibinfo{person}{Yuanhang Zhou},
  \bibinfo{person}{Wayne~Xin Zhao}, \bibinfo{person}{Xiaoke Wang}, {and}
  \bibinfo{person}{Ji-Rong Wen}.} \bibinfo{year}{2020}\natexlab{b}.
\newblock \showarticletitle{Towards topic-guided conversational recommender
  system}.
\newblock \bibinfo{journal}{\emph{arXiv preprint arXiv:2010.04125}}
  (\bibinfo{year}{2020}).
\newblock


\bibitem[\protect\citeauthoryear{Zou, Chen, and Kanoulas}{Zou
  et~al\mbox{.}}{2020}]%
        {zou2020towards}
\bibfield{author}{\bibinfo{person}{Jie Zou}, \bibinfo{person}{Yifan Chen},
  {and} \bibinfo{person}{Evangelos Kanoulas}.} \bibinfo{year}{2020}\natexlab{}.
\newblock \showarticletitle{Towards question-based recommender systems}.
\newblock \bibinfo{journal}{\emph{arXiv preprint arXiv:2005.14255}}
  (\bibinfo{year}{2020}).
\newblock


\end{thebibliography}

\appendix
\section{Appendix}
\subsection{TMDKG Construction}
\label{TMDKG Appendix}
To better understand speaker's intentions and avoid noisy information in the existing open-domain knowledge graph, we build a knowledge graph in the movie domain, which is called TMDKG. TMDKG contains 6 types of entities and 15 types of relations. Table \ref{TMDKG} shows the statistics of the TMDKG.

\textbf{Data Source} We collect the movie information from The Movie Database (TMDb), which is a community built movie and TV database. For each movie or TV show, TMDb contains related information like plots, actors, reviews, and etc.

\textbf{Information Collection} The movie entities in the TMDKG are the same as the movies mentioned in REDIAL \cite{li2018towards}, a large conversational recommendation dataset in the movie domain. We use "movie name" and "released year" as keywords to search on TMDb and keep movie genres, movie cast, movie crew, movie abstract plots, movie production companies, and department of the crew to build the TMDKG. Among the 6924 movies in REDIAL, we can match 6662 (96.2\%) movies in TMDb.

\textbf{Data Processing} We first use the Named Entity Recognition tool of spaCy\footnote{\url{https://spacy.io/}} to extract the keywords from the movie abstract plots. Then we filter out the keywords, cast, production companies that occur less than 4 times. We keep those occurring at least 10 times for the crew and all for genres. In this way, we get all the nodes in TMDKG, shown in Table \ref{TMDKG:a}. For the relationship in TMDKG, movie genres, movie cast, keywords, movie production company are viewed as 4 different relations with the movie. As the crew of movies may be from different departments, we view them as different relations with movies according to the departments they belong to. The statistics of edges of TMDKG is shown in Table \ref{TMDKG:b}.
\begin{table}[H]
\caption{Statistics of The Movie Domain Knowledge Graph (TMDKG)}
\centering
\begin{subtable}[t]{\columnwidth}
    \centering
    \begin{tabular}{|c|c|}
    \hline
        \textbf{Node Type} & \textbf{Node Number}\\
    \hline
        Genre & 19 \\
    \hline
        Movie & 6924\\
    \hline
        Cast & 1861\\
    \hline
        Crew & 3523 \\
    \hline
        Keywords & 2791 \\
    \hline
        Production & 704 \\
    \hline
        All & 15822 \\
    \hline
    \end{tabular}
    \caption{Node type and number}
    \label{TMDKG:a}
\end{subtable}
\quad
\begin{subtable}[t]{\columnwidth}
    \centering
    \begin{tabular}{|c|c|}
    \hline
        \textbf{Edge Type} & \textbf{Edge Number}\\
    \hline
        Movie-Genre & 15710 \\
    \hline
        Movie-Keyword & 37702 \\
    \hline
        Movie-Cast & 15894\\
    \hline
        Movie-Crew & 3784\\
    \hline
        Movie-Production Department & 13485\\
    \hline
        Movie-Sound & 18489\\
    \hline
        Movie-Editing & 3425\\
    \hline
        Movie-Directing & 3367\\
    \hline
        Movie-Writing  & 2915\\
    \hline
        Movie-Art & 6117 \\
    \hline
        Movie-Costume \& Make-up & 4642 \\
    \hline
        Movie-Camera & 5301 \\
    \hline
        Movie-Visual Effect & 2044 \\
    \hline
        Movie-Lighting & 445 \\
    \hline
        Movie-Production Company & 10268 \\
    \hline
        ALL & 141564 \\
    \hline
    \end{tabular}
    \caption{Edge type and number}
    \label{TMDKG:b}
\end{subtable}
\label{TMDKG}
\end{table}

\end{document}